\documentclass{article}

\usepackage{iclr2027_conference,times}

\usepackage{amsmath,amsfonts,bm}

\def\eqref#1{equation~\ref{#1}}

\def\1{\bm{1}}

\DeclareMathAlphabet{\mathsfit}{\encodingdefault}{\sfdefault}{m}{sl}
\SetMathAlphabet{\mathsfit}{bold}{\encodingdefault}{\sfdefault}{bx}{n}

\usepackage{booktabs}
\usepackage{colortbl}
\usepackage{graphicx}
\usepackage{hyperref}
\usepackage{cleveref}
\usepackage{microtype}
\usepackage{multirow}
\usepackage{placeins}
\usepackage{listings}
\usepackage{pifont}
\usepackage{subcaption}
\usepackage{url}
\usepackage{tikz}
\usetikzlibrary{arrows.meta}

\lstdefinestyle{annotationprompt}{
  basicstyle=\ttfamily\fontsize{6.8pt}{7.8pt}\selectfont,
  breaklines=true,
  breakatwhitespace=true,
  columns=fullflexible,
  keepspaces=true,
  showstringspaces=false,
  frame=l,
  framerule=0.3pt,
  framesep=0.6em,
  xleftmargin=1em,
  xrightmargin=0.5em,
  aboveskip=0.5em,
  belowskip=0.5em
}

\hypersetup{
  colorlinks=true,
  citecolor=blue,
  linkcolor=blue,
  urlcolor=black
}

\crefname{figure}{Fig.}{Figs.}
\Crefname{figure}{Fig.}{Figs.}
\crefname{table}{Tab.}{Tabs.}
\Crefname{table}{Tab.}{Tabs.}
\crefname{section}{\S}{\S\S}
\Crefname{section}{\S}{\S\S}
\crefname{subsection}{\S}{\S\S}
\Crefname{subsection}{\S}{\S\S}
\crefname{subsubsection}{\S}{\S\S}
\Crefname{subsubsection}{\S}{\S\S}
\crefname{appendix}{Appx.}{Appxs.}
\Crefname{appendix}{Appx.}{Appxs.}

\title{VidaForge: Open Research Infrastructure\\for Video Pretraining Data Recipes}

\author{%
  \textbf{Yan Ma}$^{3,5}$ \quad \textbf{Jiadi Su}$^{3,5}$ \quad \textbf{Zhulin Hu}$^{1,5}$ \quad \textbf{Ethan Chern}$^{1,2,5}$\\
  \textbf{Linhao Zhang}$^{4,5}$ \quad \textbf{Tiantian Mi}$^{2,3,5}$ \quad \textbf{Pengfei Liu}$^{1,2,5}$\\[4pt]
  \normalfont\small $^{1}$Shanghai Jiao Tong University \quad $^{2}$Shanghai Innovation Institute\\
  \normalfont\small $^{3}$Fudan University \quad $^{4}$Shanghai University\\
  \normalfont\small $^{5}$Generative Artificial Intelligence Research Lab (GAIR)
}

\iclrfinalcopy

\begin{document}

\maketitle
\lhead{Preprint}

\begin{abstract}
Video foundation models increasingly rely on large-scale pretraining data, yet their data pipelines remain largely closed and difficult to inspect or reuse. Researchers studying video data recipes often need to build substantial infrastructure before testing even a focused hypothesis. We present \textsc{VidaForge}, an open research infrastructure that connects video processing to model training through an executable five-stage workflow. VidaForge enables researchers to modify recipes, reuse processing results, and trace how each training sample was produced while maintaining throughput comparable to industrial data-curation frameworks. Using VidaForge, we study whether a fixed early-pretraining budget is better spent on a broader video pool or repeated training on a smaller, curated subset. Across video generation and self-supervised video representation learning, the broader pool yields higher mean downstream scores. The recipes with the lowest validation loss do not achieve the best downstream performance. We further release \textsc{VidaForge-3M}, containing 3.14 million scene-level clips totaling 6,475 hours, with fine-grained annotations and curation signals for video data-recipe research.

\end{abstract}

\suppressfloats[t]
\section{Introduction}
\label{sec:introduction}

\begin{figure}[tb]
  \centering
  \includegraphics[width=0.80\linewidth]{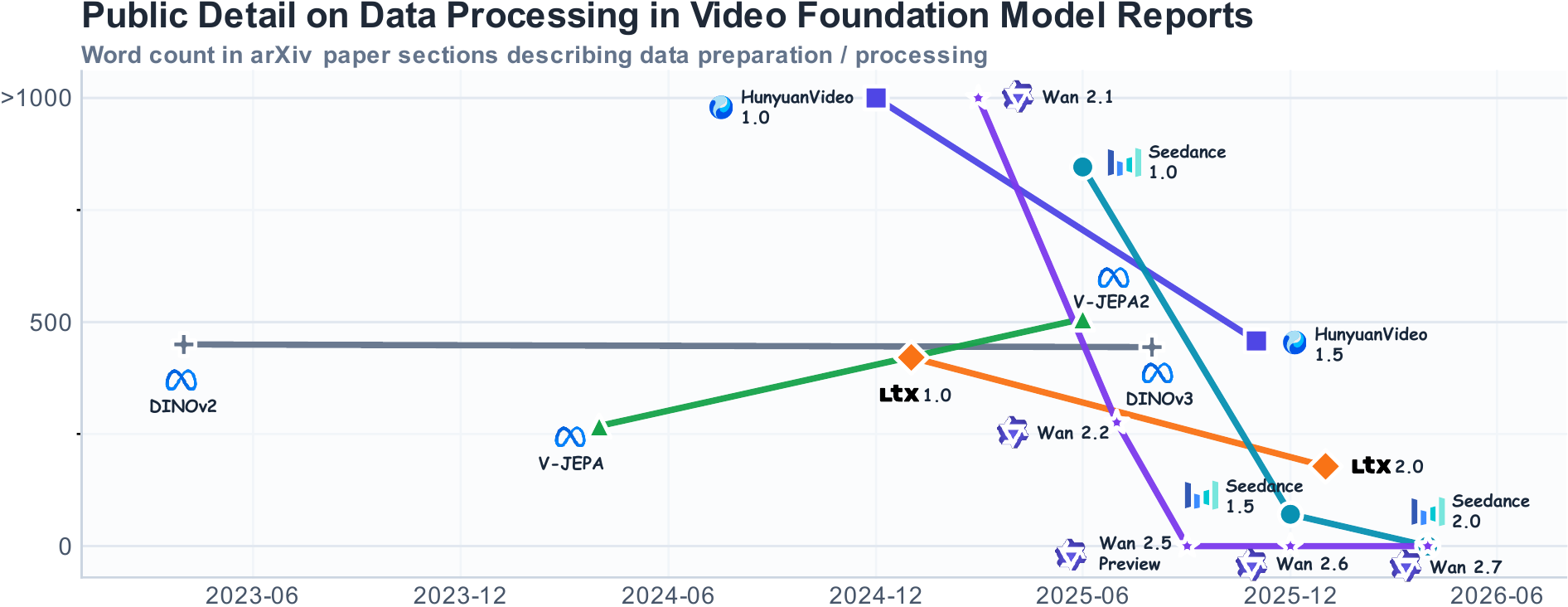}
  \caption{Publicly reported detail on data preparation and processing for recent video and visual foundation models, measured by approximate word count.}
  \label{fig:public-data-detail}
  \vspace{-10px}
\end{figure}

Video foundation models advance rapidly across generation and representation learning~\citep{wanteam2025wan,gao2025seedance,kong2024hunyuanvideo,assran2025vjepa2}, with large-scale pretraining data central to this progress~\citep{chen2026scaling,zheng2026moving,wu2026motion}.  Yet the data pipelines behind leading models remain largely closed.  As \cref{fig:public-data-detail} shows, technical reports for many leading video foundation models devote little or no space to their pretraining data processing.  They rarely report the concrete choices behind filtering, deduplication, and captioning or provide controlled training ablations showing which recipe decisions matter and under which pretraining objectives.  Consequently, the science of video pretraining data remains largely confined to a few frontier labs.

Studying data recipes requires constructing alternative training datasets and evaluating their effects under controlled training conditions~\citep{gadre2023datacomp,zheng2026moving}.  For most academic researchers, this begins with a substantial infrastructure barrier: building a pipeline from heterogeneous raw videos to the exact dataset format required by a training repository.  To compare recipes, the pipeline must preserve intermediate results for reuse, retain sample-level processing decisions, and carry the resulting dataset variants into training and evaluation.  These records connect each training sample to the choices that produced it.  Building and validating this path makes infrastructure an entry cost of video data research.
Video data infrastructure should also scale the ability to study data recipes within an academic lab's resource budget.  This requires processing results reusable across experiments: saved video scores reusable under different selection thresholds, excluded clips that remain available for comparison, and dataset variants connected to model training and evaluation.

We present \textsc{VidaForge}, an open research infrastructure designed to make video data recipes easy to modify, compare, and evaluate through model training.  As illustrated in \cref{fig:vidaforge-overview}, VidaForge turns raw videos into annotated clips and model-ready training datasets through a five-stage decision chain: ingestion, segmentation, selection, annotation, and packaging.  It saves stage outputs, recipe parameters, and sample-level decisions so that recipe changes can reuse unaffected processing results and each training sample's history can be reconstructed.  These records support controlled comparisons between recipe variants.
Using VidaForge, we investigate a practical question for video data-recipe
research under academic compute constraints: with a fixed early-pretraining
budget, is it better to train on a
broader video pool or revisit a smaller, quality-selected subset?
We compare four recipes for video generation~\citep{wanteam2025wan} and
self-supervised video representation learning~\citep{assran2025vjepa2}.
Under the same early-pretraining budget, training on the broader video pool
achieves higher mean downstream scores than reusing the smaller, curated
subset (\cref{sec:evaluation}).
Loss-based evaluation favors different recipes, highlighting the importance
of following data decisions through model evaluation.

We make three contributions. First, we introduce and open-source \textsc{VidaForge}, an end-to-end infrastructure that makes video pretraining data recipes executable and traceable, accompanied by runnable recipes, documentation, and a complete video walkthrough covering installation and usage. VidaForge achieves throughput comparable to industrial data-curation frameworks (\cref{sec:system-evaluation}). Second, we use VidaForge to study coverage--quality choices under limited early-pretraining budgets, comparing full-pool training with selected, random, and rejected subsets across generative and representation learning. Third, we release \textsc{VidaForge-3M}, an open 3.14-million-clip, 6,475-hour dataset with fine-grained annotations and curation signals for video data recipe research.

\section{Related Work}
\label{sec:related-work}
\label{app:related-work}

Technical reports on video and visual foundation models~\citep{gao2025seedance,seedance2026seedance,wanteam2025wan,hong2022cogvideo,yang2024cogvideox,opensora,opensora2,kong2024hunyuanvideo,hunyuanvideo2025,HaCohen2024LTXVideo,hacohen2026ltx,assran2025vjepa2,simeoni2025dinov3}, such as Seedance, Wan, and V-JEPA~2, describe aspects of training-data preparation, including filtering and captioning.  These reports, however, do not consistently specify the full data recipe and its implementation.  Researchers must first build the data-processing infrastructure needed to construct datasets with different recipes.  They can then study how these choices affect pretraining by holding the model and training budget fixed.

Alibaba's Data-Juicer~\citep{djv2} provides a broad library of processing operators spanning text, image, audio, and video.  Data-Juicer Sandbox~\citep{chen2025datajuicersandbox} studies how text, image, and video processing operators affect downstream model performance.  NVIDIA's Curator frameworks~\citep{nemo_curator,cosmos-curator} support large-scale data curation across modalities.  Beyond scaling processing throughput, operator libraries, and modality coverage, VidaForge aims to make traceable video data-recipe experimentation practical.  Researchers can compare recipe variants from a shared processed pool through both generative and self-supervised pretraining.  This design allows processing results to support subsequent experiments beyond the initial dataset construction.  \Cref{tab:framework-positioning} compares the capabilities supporting this workflow.

Large-scale video datasets such as InternVid~\citep{wang2024internvid}, Panda-70M~\citep{chen2024panda}, OpenVid-1M~\citep{nan2025openvid}, and Sekai~\citep{li2026sekai} provide curated corpora for video model training, with their public releases centered on the resulting data artifact.  A small but growing body of work further isolates specific factors in training data, including caption quality, data composition, and sample-level influence~\citep{chen2026scaling,zheng2026moving,wu2026motion}.  VidaForge provides reusable infrastructure for extending such studies across multiple pipeline stages and carrying traceable dataset variants into pretraining experiments.

\section{VidaForge}
\label{sec:vidaforge}

\begin{figure}[t]
    \centering
    \includegraphics[width=0.95\linewidth]{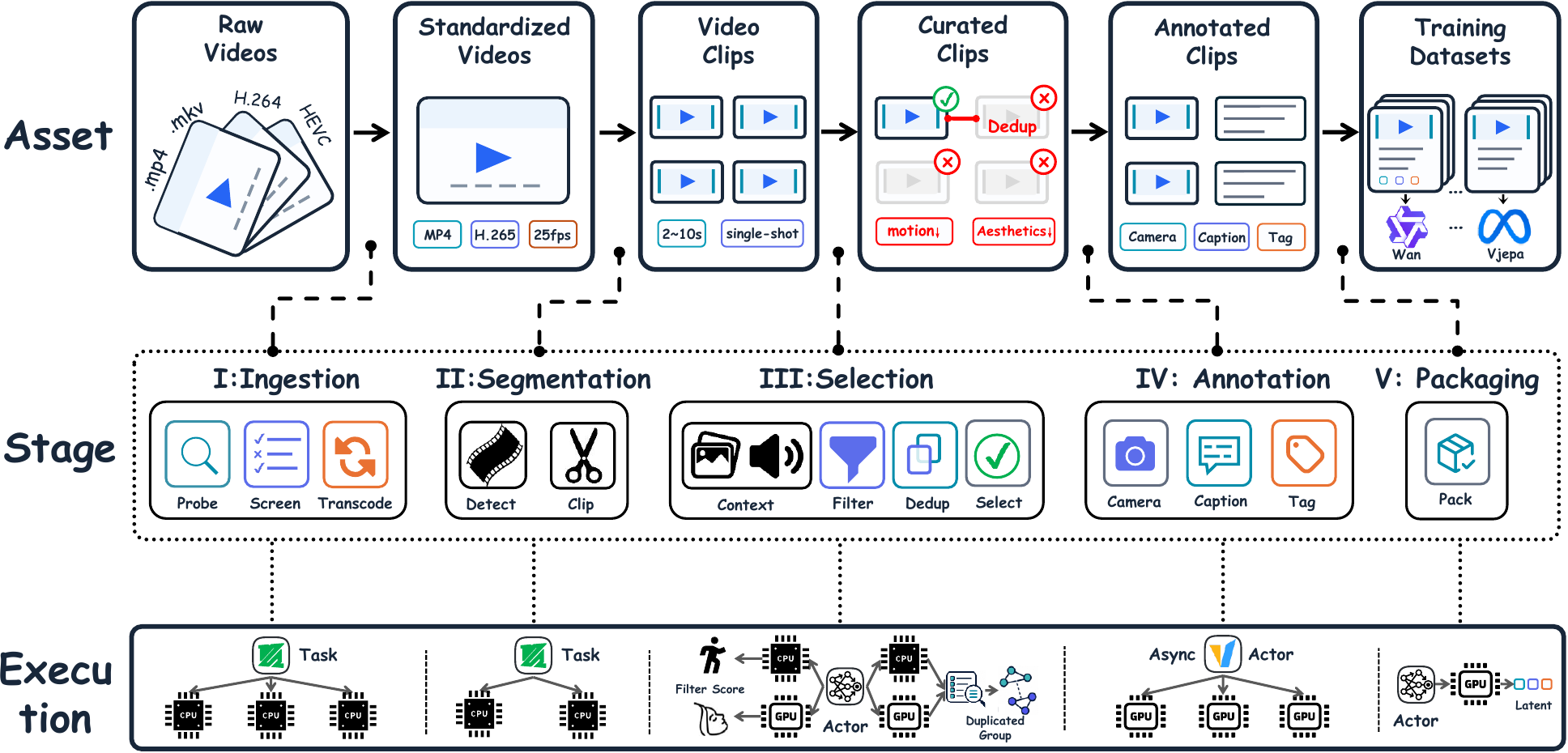}
    \caption{\textbf{VidaForge overview.} The top row follows six data states from raw videos $\mathcal{D}_0$ to target-specific training datasets $\mathcal{D}_5$.  Five stages implement the transitions between these states, while the bottom row shows the execution patterns used by their processing steps.}
    \label{fig:vidaforge-overview}
    \vspace{-10pt}
\end{figure}

VidaForge treats support for data-recipe experiments as a core design requirement: each processing run produces a training dataset and reusable results for subsequent recipe comparisons.  Let $\mathcal{D}_0$ denote a fixed snapshot of raw videos.  As illustrated in \cref{fig:vidaforge-overview}, VidaForge represents a data recipe $\mathcal{R}$ as a chain of five stage-level transformations:
\begin{equation}
    \mathcal{D}_0
    \xrightarrow[\text{\scriptsize Ingestion}]{\text{\scriptsize Stage I}}
    \mathcal{D}_1
    \xrightarrow[\text{\scriptsize Segmentation}]{\text{\scriptsize Stage II}}
    \mathcal{D}_2
    \xrightarrow[\text{\scriptsize Selection}]{\text{\scriptsize Stage III}}
    \mathcal{D}_3
    \xrightarrow[\text{\scriptsize Annotation}]{\text{\scriptsize Stage IV}}
    \mathcal{D}_4
    \xrightarrow[\text{\scriptsize Packaging}]{\text{\scriptsize Stage V}}
    \mathcal{D}_5.
    \label{eq:recipe-chain}
\end{equation}
The complete chain defines $\mathcal{D}_5=\mathcal{R}(\mathcal{D}_0)$.  Its intermediate states contain standardized videos $\mathcal{D}_1$, video clips $\mathcal{D}_2$, clips with selection outcomes $\mathcal{D}_3$, and annotated clips $\mathcal{D}_4$; $\mathcal{D}_5$ is the target-specific training dataset.  Each state includes the video assets and accumulated processing records.  In particular, $\mathcal{D}_3$ retains both accepted and rejected clips with their scores and decisions, allowing different training subsets to be constructed from the same processed pool.  Each arrow is implemented by a sequence of configured processing steps.  Changing a step's decision defines a recipe variant $\mathcal{R}'$ and produces an alternative dataset $\mathcal{D}'_5=\mathcal{R}'(\mathcal{D}_0)$.
VidaForge executes such recipes while preserving the data and processing record produced by every step, allowing each final sample to be traced through the recipe that produced it.

\paragraph{Executing a recipe.}
Executing a recipe turns its processing choices into a training dataset while recording what happens to each video and clip.  The top and middle rows of \cref{fig:vidaforge-overview} show the data transformations and their processing steps.
Raw videos differ in format and media properties, so Ingestion prepares standardized inputs. Probe records basic media properties, such as duration, resolution, frame rate, and codec. Screen applies low-cost checks to exclude videos with insufficient resolution, excessive duration, or frame rates below a specified threshold. Transcode converts accepted videos to H.265/MP4 and downscales high-resolution inputs.
A video can contain multiple scenes, so Segmentation creates scene-level units for scoring, annotation, and training-data selection: Detect locates scene boundaries using PySceneDetect~\citep{pyscenedetect} or TransNetV2~\citep{soucek2020transnetv2}, and Clip extracts the corresponding clips with FFmpeg.
Selection separates measuring clips from deciding which enter training, so the same measurements can support different selection rules. Context extracts frames for reuse by filtering, Camera, Caption, and Tag, and makes audio available to steps that require it. We implement four representative filters. Optical multiplies exposure and contrast scores aggregated from sampled frames. Motion measures frame-to-frame changes using FFmpeg's VMAFMOTION~\citep{netflixvmaf}, with a penalty for frequent scene cuts. Aesthetics uses Aesthetic Predictor V2.5~\citep{aestheticpredictorv25}, a SigLIP-based model with a learned scoring head. Visible text uses PP-OCRv5~\citep{cui2026ppocrv5} to estimate text coverage in sampled frames, with higher scores indicating less text. Dedup identifies perceptually similar clips through Hamming-distance matching of frame hashes and semantically similar clips through cosine similarity of video embeddings. Select combines these saved signals to record which clips pass, retaining both accepted and rejected clip records.
Camera motion contributes to the apparent motion in a video alongside subject movement. Annotation therefore includes a separate Camera step, whose schema adapts the CameraBench taxonomy~\citep{lin2025camerabench} to describe motion type, steadiness, rotation, translation, and zoom. Caption uses a vision-language model to read sampled frames with Camera labels as auxiliary context and generate four description levels, from a short summary to a detailed account. These levels allow researchers to vary annotation detail while keeping the video set fixed. Tag independently assigns structured labels from the frames, covering dimensions such as scene, subjects, actions, and visual style; \cref{app:annotation-schemas,app:annotation-prompts} provide their schemas and prompts.
Finally, Packaging supports two representative video-learning objectives: text-conditioned video generation and self-supervised video representation learning. For generation, it computes video latents and text embeddings for Wan training through NeMo-AutoModel. For representation learning, it prepares video lists for the official V-JEPA~2 training code. Both training formats retain clip identifiers for tracing processing history. \Cref{app:stage-packaging} details these conversions.

\paragraph{Varying a recipe.}
To study a processing choice, a recipe variant reuses saved results from before the changed step and reruns the later steps that depend on it.  Changing segmentation changes where videos are split; changing selection changes which clips enter training; and changing Camera, Caption, or Tag changes the descriptions or labels assigned to each clip.
Researchers can change selection thresholds or duplicate-retention rules using saved scores and duplicate relations, without repeating scoring or matching. Previously excluded clips remain available for reselection and training comparisons.
Saved clip identities and processing records let researchers compare which clips were selected and how their annotations changed across recipes. When segmentation changes, clips remain traceable to their source videos and temporal intervals.

\paragraph{Scaling a recipe.}
To make these studies practical on large video pools, VidaForge distributes media processing and model inference across CPU and GPU resources.
The bottom row of \cref{fig:vidaforge-overview} summarizes how VidaForge parallelizes each step according to the work it performs.  Probe, Transcode, and Clip use FFmpeg/ffprobe for media inspection and processing, with Ray distributing videos across CPU workers for large-scale parallel execution~\citep{moritz2018ray}.  GPU-based filters and semantic embedding models run on GPU workers, each loading its model once and processing successive batches of clips.  Camera, Caption, and Tag send concurrent requests to a pool of inference servers.  Deduplication must compare clips across the full dataset and therefore runs in two phases.  Feature-extraction workers first compute visual hashes or semantic embeddings and save them in shards.  For implementation simplicity, matching workers each load the full index and process separate query subsets in parallel, avoiding cross-shard query routing and result merging.  Matched pairs form candidate duplicate groups. For semantic deduplication, a clip is removed only if its cosine similarity to an initially retained clip meets the configured threshold.  Each execution pattern saves step outputs and processing records, allowing completed work to be reused when a recipe resumes.  \Cref{app:detailed-pipeline} describes every stage and step in detail.

\section{Coverage or Quality? Studying Video Data Recipes with VidaForge}
\label{sec:evaluation}

Given the cost of full-scale video pretraining, we focus on early pretraining
within academic compute budgets. Using VidaForge, we ask: \emph{Under a fixed
training budget, is it more effective to train on a broader video pool or
repeatedly train on a smaller, curated subset?} We construct four recipes from
the same processed video pool by varying quality filtering, deduplication, and
sampling while holding the remaining processing choices fixed. Within each
model family, we match training steps and the number of processed clips.
We train Wan~2.1 for video generation~\citep{wanteam2025wan} and V-JEPA~2-1B for
self-supervised video representation learning~\citep{assran2025vjepa2} from
scratch, evaluating downstream performance with VBench and frozen-probe action
recognition, respectively.

\begin{figure}[t]
  \centering
  \includegraphics[width=0.85\linewidth]{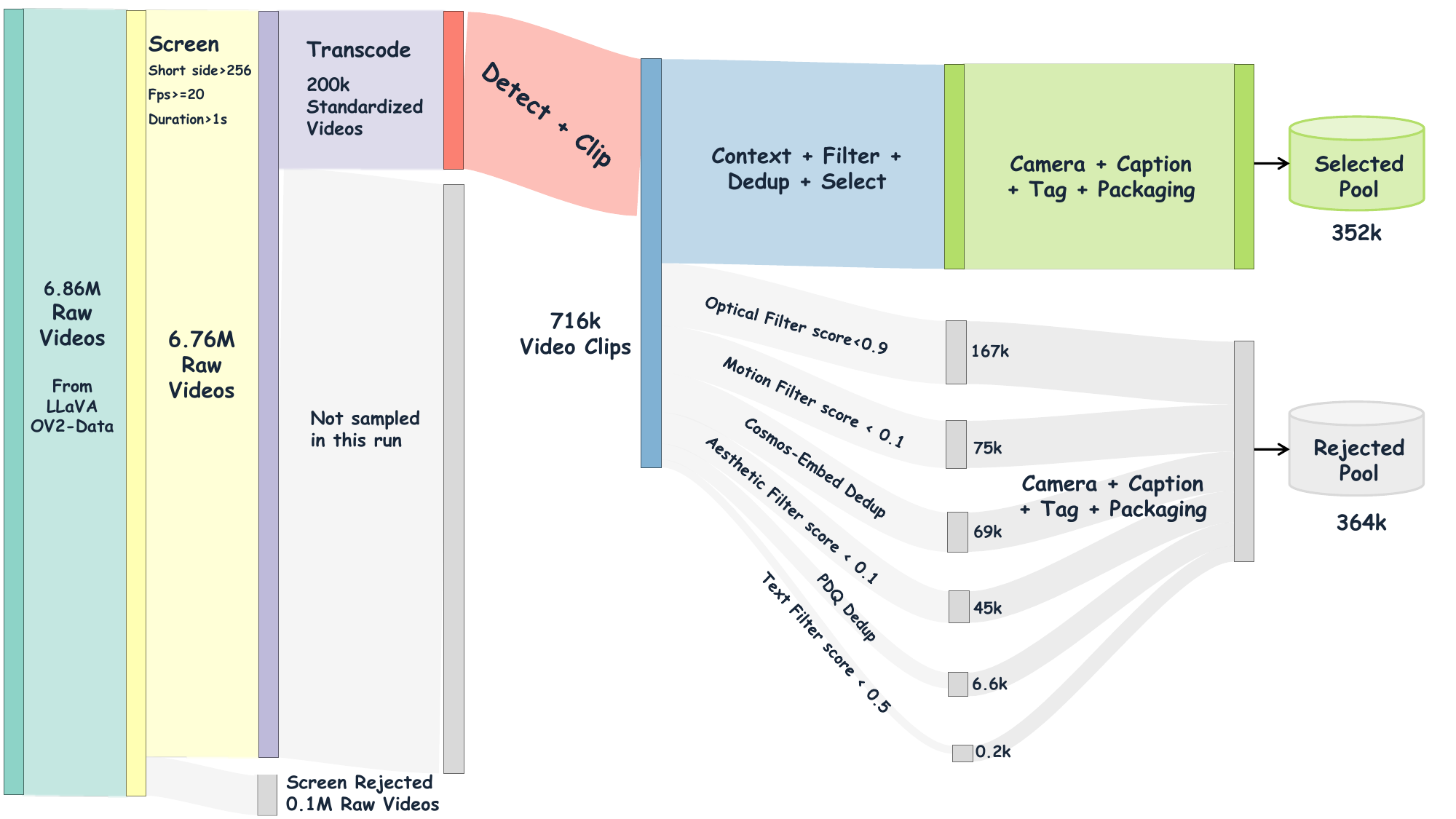}
  \caption{VidaForge data flow for the coverage--quality study.  Blue marks
  clips admitted by Selection; gray branches show clips rejected by individual
  filters or deduplication thresholds. Each rejected clip is counted under the
  first rule it fails. The two outcomes form the Selected and
  Rejected pools.}
  \label{fig:recipe-flow}
  \vspace{-4pt}
\end{figure}

\paragraph{From a Video Pool to Recipe Variants.}
\Cref{fig:recipe-flow} traces one execution over the video portion of
LLaVA-OneVision-2-Data~\citep{an2026llavaonevision2}.  Starting from 200k
standardized videos, the fixed Ingestion and Segmentation path produces 716k
scene-level clips. We expand processing to 800k source videos from the same
collection to construct \textsc{VidaForge-3M}; \cref{app:vidaforge3m} provides
release statistics. We evaluate quality filtering and deduplication together
as a complete data-selection recipe.
The optical, motion, aesthetic, and visible-text scores range from 0 to 1,
with higher values preferred by this recipe; a higher visible-text score
indicates less detected text area.  Clips must score at least
0.9, 0.1, 0.1, and 0.5, respectively, to pass the four filters.
We form candidate duplicate groups using PDQ and Cosmos-Embed matches.
Among clips passing the quality filters, PDQ retains the member with the
highest sum of the four scores in each group. We rank the remaining clips
within each Cosmos-Embed group by the same score and initially retain the
top 20\%, rounded up and capped at 20 clips, with at least one retained.
Each remaining clip is excluded only if its cosine similarity to an initially
retained clip is at least 0.95; clips without such a match are additionally retained.
Clips that pass all
filtering and deduplication rules form the Selected pool; clips excluded by
any rule form the Rejected pool. This policy produces 352k Selected and 364k Rejected
clips.  Both pools retain their measurements and decision records;
\cref{app:stage-selection} details duplicate matching and the full selection policy,
with visual examples in \cref{app:selection-examples}.
For annotation, Camera uses Gemma-4-E4B-it~\citep{google2026gemma4e4bit},
while Caption and Tag use Qwen3.6-27B-FP8~\citep{qwen2026qwen36fp8}.
Wan uses level-3 captions, the most detailed of the four caption levels
described in \cref{app:caption-schema}.  \Cref{app:annotation-examples} shows
annotations for both a selected and a rejected clip.

\subsection{Training and Validation Sets}
\label{sec:training-comparison}

\textbf{Dataset construction.} We compare four recipes under the same budget
of approximately 580k processed clips: Mixed uses approximately 580k distinct
clips for 1 epoch, while Selected, Rejected, and Random each use approximately
290k distinct clips for 2 epochs. To construct these datasets, we first reserve
three 10k-clip validation sets from the Selected and Rejected pools. The
Selected and Rejected validation sets draw 10k clips from their corresponding
pools; the Mixed validation set draws 5k clips from each.  Every clip sharing a
parent \texttt{video\_id} with any validation clip is then excluded from all
training recipes. Selected and Rejected draw from their corresponding pools
after validation isolation. Mixed serves as the baseline, combining
approximately 290k clips from each pool.
Random uses a random subset of Mixed containing $\approx$145k clips
from each pool. Each subset is fixed within a run. Selected versus Random compares
selection strategies at the same dataset size and epoch count; Mixed versus
Random compares the full pool with repeated use of a smaller random subset.
Rejected tests the training value of clips excluded by the selection policy.

\textbf{From-Scratch Pretraining.} The same recipe variants are evaluated in two models trained from scratch:
Wan~2.1-1.3B for generation and V-JEPA~2-1B for self-supervised representation
learning.  Within each model family, architecture, optimization, training steps,
and evaluation protocols are held fixed across recipes. We conduct 24 from-scratch
pretraining runs: 4 recipes $\times$ 3 runs $\times$ 2 model families.
Random uses an independently sampled subset in each run, so its reported
variation includes both sampling and training randomness.
For each run, we report validation loss on 3 shared validation sets at
7 checkpoints for Wan and 20 for V-JEPA~2. VBench evaluates 3 checkpoints
per Wan run, with 3 videos per prompt.
SSv2 evaluates 5 checkpoints per V-JEPA~2 run, while
Kinetics-400 evaluates the final checkpoint from one run per recipe.
Each SSv2/Kinetics-400 evaluation trains 3 probes with different learning
rates for 20 epochs. \Cref{tab:experiment-scale} summarizes the evaluation
counts. Dataset construction
and training protocols are detailed in \cref{app:recipe_interventions} and
\crefrange{app:training-controls}{app:vjepa-details}.

\begin{figure}[t]
  \centering
  \begin{subfigure}[c]{0.532\linewidth}
    \centering
    \includegraphics[width=\linewidth]{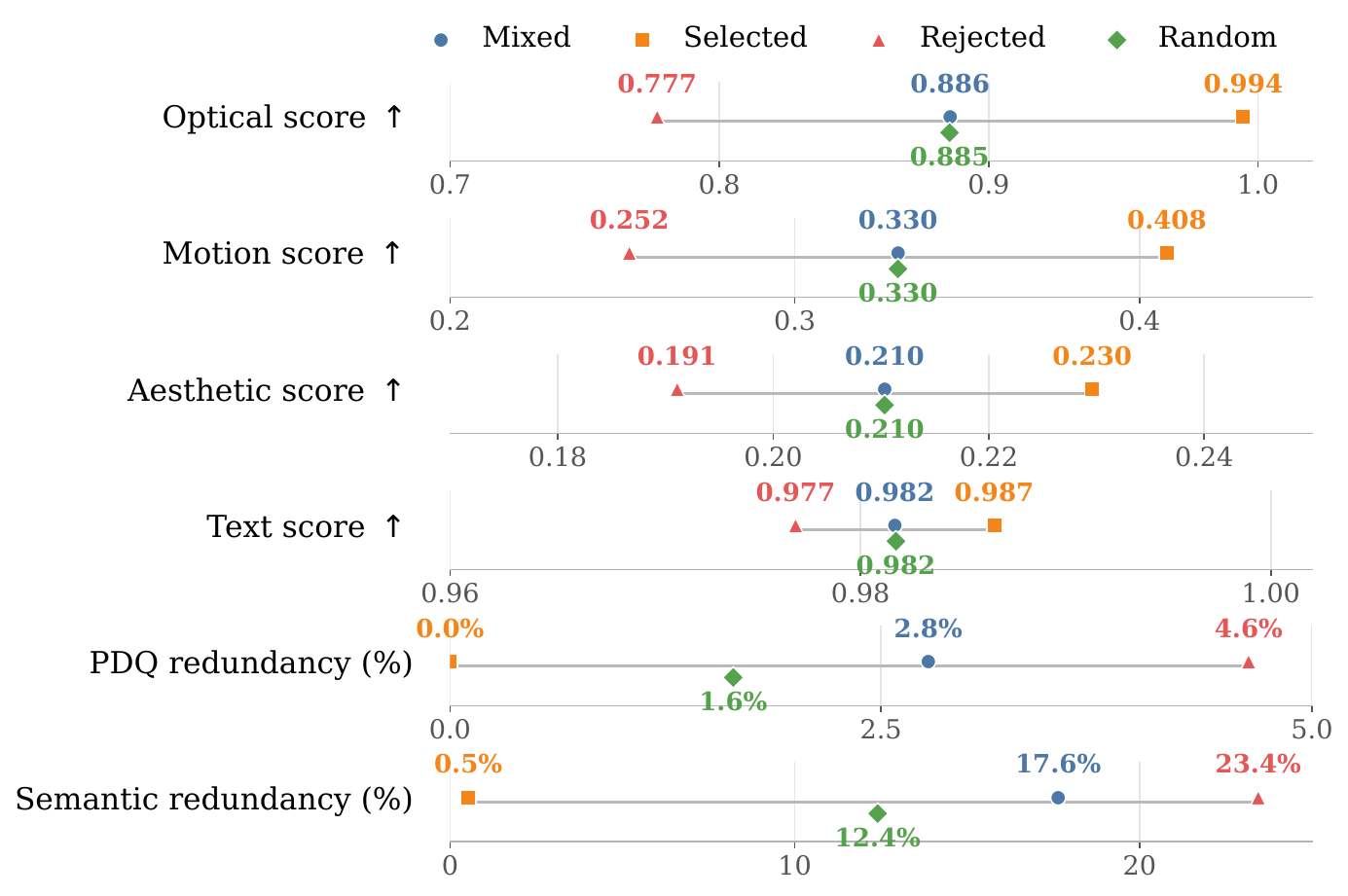}
    \caption{Quality scores and redundancy.}
    \label{fig:recipe-quality}
  \end{subfigure}\hspace{0.0285\linewidth}
  \begin{subfigure}[c]{0.3895\linewidth}
    \centering
    \includegraphics[width=\linewidth]{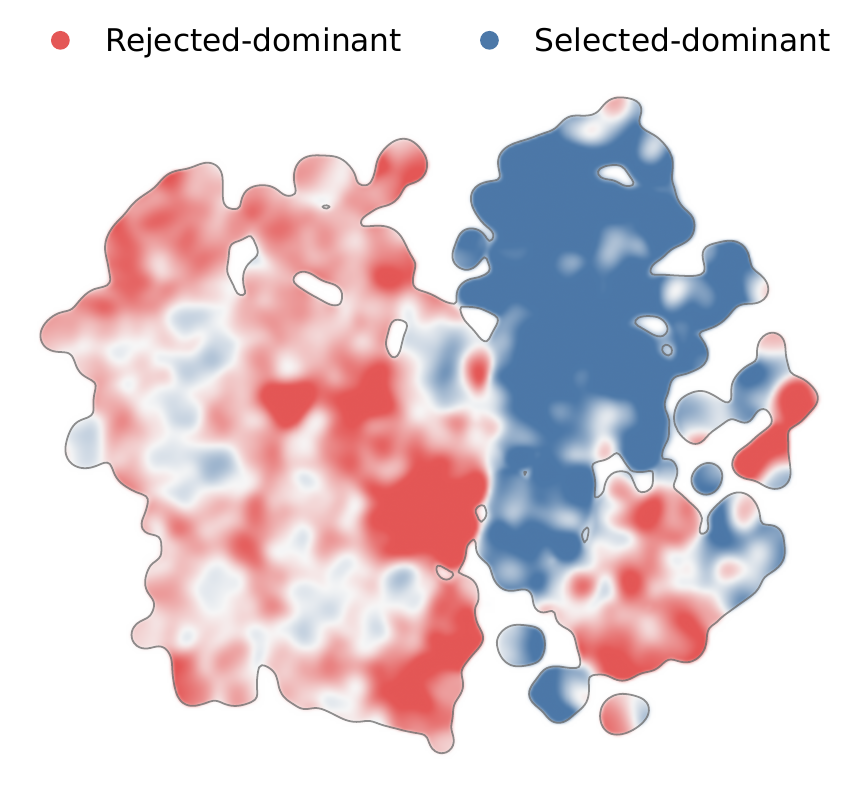}
    \caption{Coverage in video embedding space.}
    \label{fig:recipe-semantic-coverage}
  \end{subfigure}
  \caption{Quality and coverage of the training recipes. Left: average
  optical, motion, aesthetic, and visible-text scores, together with the
  within-dataset redundancy measured using PDQ~\citep{meta2019pdq} and
  Cosmos-Embed~\citep{nvidia2025cosmosembed1} groups, for all four recipes.
  Random values are averaged over 3 sampled subsets. Right: a
  t-SNE map of semantic video embeddings generated by Cosmos-Embed, sampled
  from the candidate pool.  Blue regions contain more Selected clips and red
  regions contain more Rejected clips; $\text{Mixed}=\text{Selected}\cup
  \text{Rejected}$ covers both.}
  \label{fig:recipe-profiles}
  \vspace{-4pt}
\end{figure}

\textbf{Quality and coverage.} Quality is characterized by the four filtering
scores and within-dataset redundancy; coverage is characterized by the
number of distinct clips and their distribution in semantic embedding space.
The three constituent datasets are
Mixed (coverage $\uparrow$, mixed
quality), Selected (coverage $\downarrow$, quality $\uparrow$), and Rejected
(coverage $\downarrow$, quality $\downarrow$).  Selected has higher optical,
motion, and aesthetic scores and lower redundancy than Rejected, with
Mixed lying between them (\cref{fig:recipe-quality}).  In the video embedding
map, Selected and Rejected concentrate in different regions, while Mixed
covers both (\cref{fig:recipe-semantic-coverage}). Random has nearly the same
mean quality scores as Mixed while using half as many clips, with approximately
equal Selected/Rejected proportions. Subsampling leaves fewer clips together
in duplicate groups, lowering Random's redundancy relative to Mixed, though
it remains higher than Selected.
Full score and duplicate-group
distributions for these three datasets appear in
\cref{fig:recipe-score-distributions,fig:recipe-duplicate-distributions}.

\subsection{Wan~2.1-1.3B: Video Generation}
\label{sec:wan-results}

We train Wan~2.1-1.3B from scratch with 3 runs per recipe on 32 H200 GPUs,
global batch size 96, and approximately 6,000 updates under the budget in
\cref{sec:training-comparison}.

\begin{table}[htbp]
  \centering
  \small
  \setlength{\tabcolsep}{3pt}
  \renewcommand{\arraystretch}{1.12}
  \caption{Final downstream scores under the fixed training budget in each
  model family. VBench and SSv2 report mean $\pm$ std over 3 runs;
  Kinetics-400 reports one run per recipe. Higher is better.}
  \label{tab:recipe-final-results}
  \begin{tabular}{@{}lccccc@{}}
    \toprule
    & \multicolumn{3}{c}{Wan~2.1-1.3B: VBench} & \multicolumn{2}{c}{V-JEPA~2-1B} \\
    \cmidrule(lr){2-4}\cmidrule(l){5-6}
    Recipe & Quality & Semantic & Total & SSv2 (\%) & Kinetics-400 (\%) \\
    \midrule
    Mixed & $\mathbf{71.65}\pm0.74$ & $\mathbf{16.60}\pm1.25$ & $\mathbf{60.64}\pm0.83$ & $\mathbf{16.256}\pm0.153$ & $\mathbf{27.218}$ \\
    Selected & $70.61\pm0.51$ & $16.28\pm0.81$ & $59.74\pm0.55$ & $15.712\pm0.382$ & $26.880$ \\
    Random & $70.25\pm0.80$ & $16.43\pm0.76$ & $59.48\pm0.67$ & $15.311\pm0.194$ & $25.519$ \\
    Rejected & $70.25\pm1.19$ & $16.08\pm0.89$ & $59.42\pm1.06$ & $15.099\pm0.338$ & $25.776$ \\
    \bottomrule
  \end{tabular}
  \par\smallskip
  {\footnotesize SSv2 and Kinetics-400 report top-1 accuracy.\par}
\end{table}

We evaluate each final checkpoint on VBench~\citep{huang2024vbench}, averaging
3 generations per prompt (\cref{tab:recipe-final-results}). At this early
stage, the model does not yet consistently generate coherent videos. We use
VBench's Quality and Semantic scores to compare the emerging visual and semantic
properties of videos generated by models trained on different recipes. At the same
dataset size and epoch count, Selected exceeds Random by 0.258 points in mean
Total. Using the full pool, Mixed exceeds Random by 1.152 points. Mixed
ranks first in mean Quality, Semantic, and Total.
Mixed achieves higher VBench Total than Selected in all 3 runs,
with gains of 0.53--1.15 points
(\cref{tab:wan-vbench-seeds}).
Mixed also leads in mean Imaging Quality, Aesthetic Quality, and Appearance
Style (\cref{fig:wan-vbench,tab:wan-vbench-dimensions} in
\cref{app:wan-details}).
VBench Total scores increase during training for all four recipes
(\cref{fig:wan-vbench-trajectories} in \cref{app:downstream-dynamics}).

\begin{figure}[t]
  \centering
  \includegraphics[width=\linewidth]{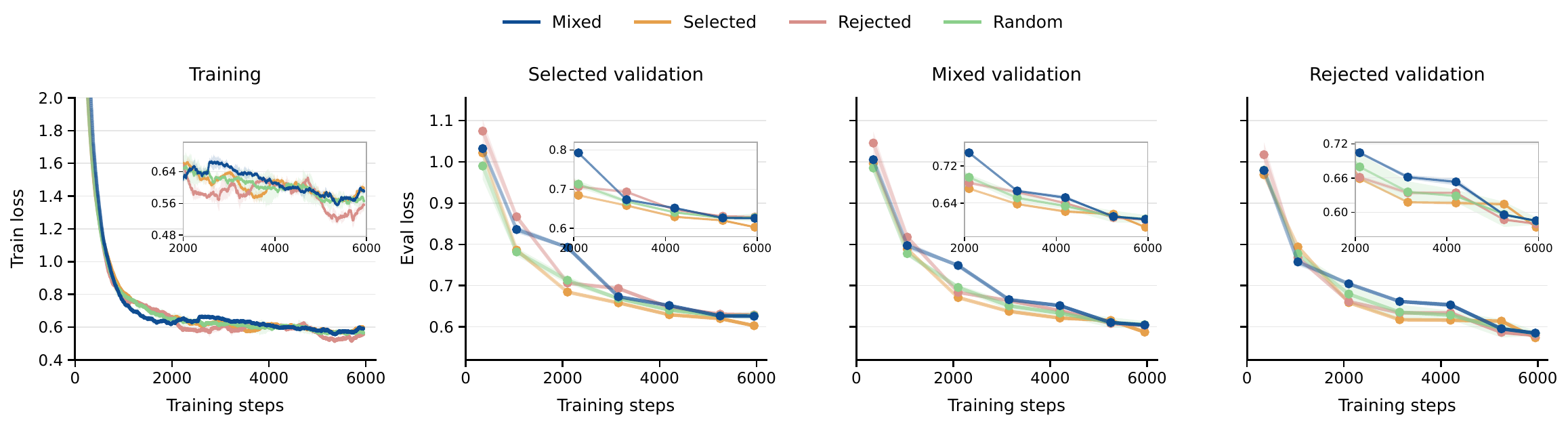}
  \caption{Wan~2.1-1.3B loss curves for all four recipes across
  3 runs (mean $\pm$ std).
  Left: training loss. Right: loss on the Selected, Mixed, and Rejected
  validation sets. Rejected has the lowest final mean training loss;
  Selected has the lowest final mean validation loss.}
  \label{fig:wan-losses}
  \vspace{-12pt}
\end{figure}

Rejected ends with the lowest mean training loss, while Selected reaches the
lowest final mean loss on all three validation sets (\cref{fig:wan-losses}).
The validation-loss ranking thus differs from VBench, which favors Mixed.
An enlarged view of the training curve appears in
\cref{fig:wan-training-loss}.

\subsection{V-JEPA~2-1B: Representation Learning}
\label{sec:vjepa-results}

We train V-JEPA~2 from scratch with 3 runs per recipe,
global batch size 256, and the budget in \cref{sec:training-comparison}.
We evaluate final encoder checkpoints using frozen attentive probes from
the official V-JEPA~2 repository on Something-Something V2
(SSv2)~\citep{goyal2017something}. At the same dataset size and epoch count,
Selected exceeds Random by 0.401 percentage points in mean accuracy
(\cref{tab:recipe-final-results}). Using the full pool, Mixed exceeds Random
by 0.945 points and has the highest mean accuracy across all four recipes.
Mixed also outperforms Selected and Rejected
under all 3 matched training seeds, with gains over Selected of
0.28--0.72 percentage points (\cref{tab:ssv2-training-seeds}).
On Kinetics-400~\citep{kay2017kinetics}, an additional single-run
frozen-probe evaluation also places Mixed first, while Selected exceeds
Random at the same dataset size (\cref{tab:recipe-final-results};
protocol in \cref{app:vjepa-details}).

\begin{figure}[t]
  \centering
  \includegraphics[width=\linewidth]{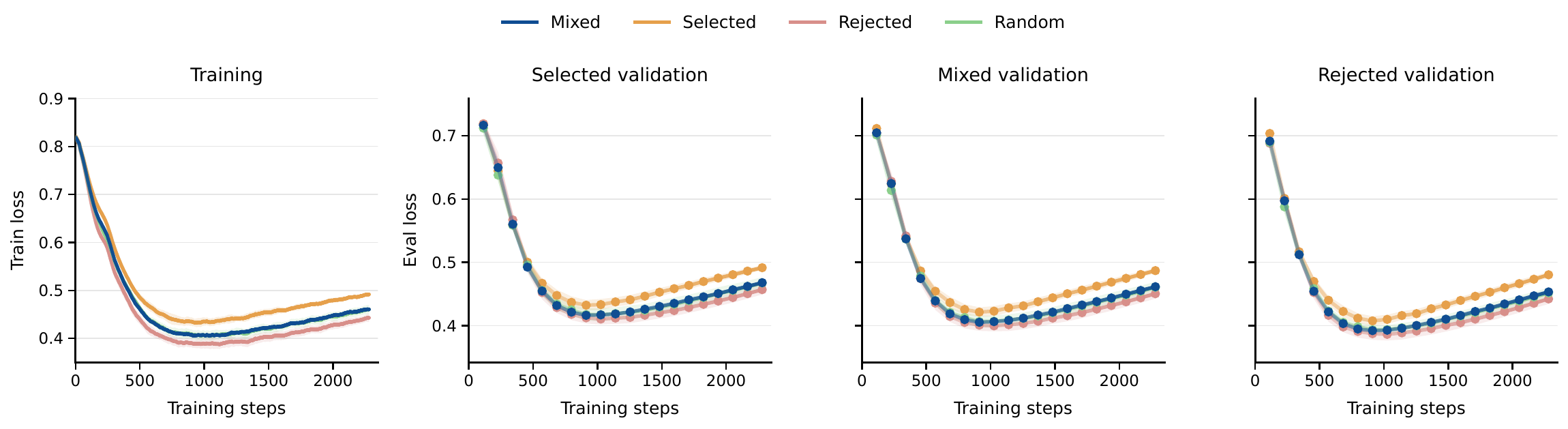}
  \caption{V-JEPA~2 loss curves across 3 runs (mean $\pm$ std).
  Left: training loss. Right: loss on the Selected, Mixed, and Rejected
  validation sets. Rejected has the lowest final mean training and validation
  losses. An enlarged training curve appears in \cref{fig:vjepa-training-loss}.}
  \label{fig:vjepa-comparison}
  \vspace{-4pt}
\end{figure}

Rejected reaches the lowest final mean loss on all three validation sets
and in training (\cref{fig:vjepa-comparison}), while Mixed has the highest
mean SSv2 accuracy. Mixed also has the lowest validation-loss variability
across seeds (\cref{app:vjepa-details}).
The trajectories for all four recipes show improving late-stage SSv2 accuracy despite
rising masked-prediction loss (\cref{fig:ssv2-trajectories});
\cref{app:downstream-dynamics} presents the downstream trajectories.

\paragraph{Main finding.}
At the same dataset size and epoch count, Selected improves mean downstream
performance over Random on both benchmarks. Under the fixed early-pretraining
budget, however, one pass over the full Mixed pool yields the highest mean
performance, ahead of two passes over the selected or random subset.
Training losses reflect each recipe's own training distribution.
On the same validation sets, however, lower pretraining loss does not identify
the recipe with the strongest downstream performance: final mean validation
loss favors Selected for Wan and Rejected for V-JEPA~2, while Mixed performs
best on VBench and SSv2.
VidaForge connects recipe choices to pretraining dynamics and downstream performance.

\section{Beyond Throughput: Supporting Video Data-Recipe Research with VidaForge}
\label{sec:system-evaluation}

The recipe study motivates two system-level questions: how efficiently can
video processing scale with additional resources, and what support do
frameworks provide for constructing and modifying data-recipe experiments?

\subsection{Competitive Throughput}
\label{sec:processing-scalability}

\begin{figure}[t]
  \centering
  \includegraphics[width=\linewidth]{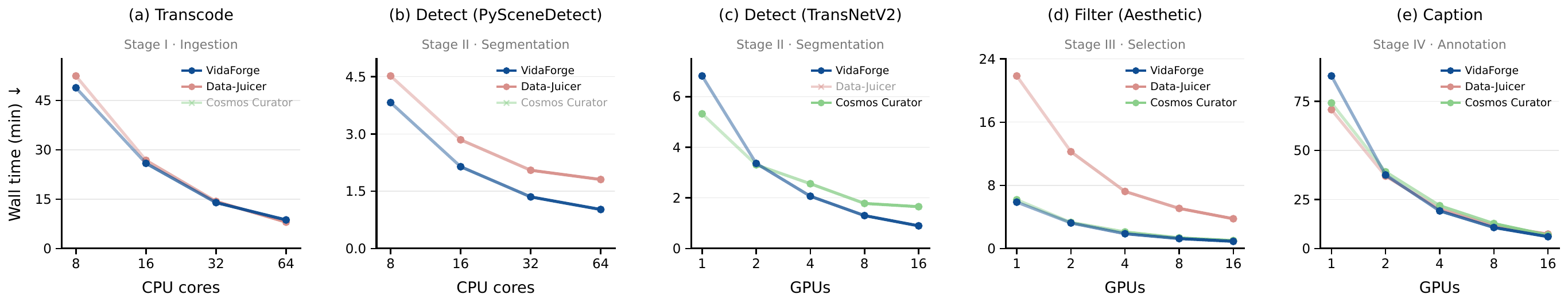}
  \caption{Processing time as CPU or GPU resources increase. Transcode and
  scene detection process 1,000 videos; aesthetic scoring and captioning
  process 10,000 clips. Lower is better. Crossed legend entries indicate
  framework--operator combinations not included in this comparison.
  Each point averages 3 runs under the configurations described in the text.}
  \label{fig:operator-scaling}
\end{figure}

We compare VidaForge with Alibaba's Data-Juicer and NVIDIA's Cosmos Curator on five video
processing operators (\cref{fig:operator-scaling}). For each operator, we use
the same input data, algorithms, model weights where applicable, preprocessing,
and output requirements across frameworks. We integrate these matched workloads
through each framework's processing interfaces and compare execution time under
the same resource allocations. Transcode and the two
scene detectors process fixed sets of 1,000 videos; aesthetic scoring and
captioning process 10,000 clips. CPU workloads use 8--64 cores, and GPU
workloads use 1--16 NVIDIA H200 GPUs. All frameworks run on nodes equipped with
Intel Xeon Platinum 8558 CPUs. Each point reports mean wall-clock time over 3 runs.
PySceneDetect and TransNetV2 are evaluated separately. Aesthetic scoring
uses Aesthetic Predictor v2.5 with SigLIP SO400M, integrated into the external
frameworks through custom operators; captioning uses Qwen3.6-27B-FP8.
Aesthetic scoring and captioning use pre-extracted frames, so their reported
times exclude frame extraction.
\Cref{app:system-execution} additionally reports processing throughput during
construction of \textsc{VidaForge-3M}.

VidaForge's processing time decreases with additional resources across all
five operators. PySceneDetect drops from 229 to 61 seconds from 8 to 64 CPU
cores, a $3.74\times$ throughput increase; at 64 cores it takes 43\% less
time than Data-Juicer. Transcode is competitive with Data-Juicer, with
Data-Juicer faster at 64 cores. At 8 GPUs, VidaForge takes 78 seconds for
TransNetV2 versus Cosmos Curator's 107 seconds, and 642 seconds for captioning
versus 697 for Data-Juicer and 763 for Cosmos Curator. These results show
competitive processing efficiency alongside VidaForge's recipe workflow.

\subsection{From Video Processing to Recipe Experiments}
\label{sec:recipe-research-support}

Supporting data-recipe research requires preserving the results of one
processing run for further experiments: reusing completed processing,
reconsidering excluded clips, and comparing variants through model training.
\Cref{tab:framework-positioning} compares how released workflows support
these needs, based on their code, configurations, and documented training
interfaces. \Cref{app:workflow-definitions} gives the definitions and scope.

\begin{table}[t]
  \centering
  \caption{Support for video processing and data-recipe research in released
  workflows. \ding{51} indicates that the full row definition is met;
  --- indicates that it is not, including partial support.
  Official extensions are included; Data-Juicer includes Sandbox~\citep{chen2025datajuicersandbox}.}
  \label{tab:framework-positioning}
  \begingroup
  \definecolor{wfprocessing}{HTML}{E8EDF2}
  \definecolor{wfprocessingbody}{HTML}{F5F7F9}
  \definecolor{wfresearch}{HTML}{DCEEEA}
  \definecolor{wfresearchbody}{HTML}{F3F8F6}
  \newcommand{\wfyes}{\ding{51}}
  \newcommand{\wfno}{---}
  \small
  \setlength{\tabcolsep}{7pt}
  \renewcommand{\arraystretch}{1.15}
  \begin{tabular}{lcccc}
    \toprule[0.6pt]
    & Data-Juicer & NeMo Curator & Cosmos Curator
      & \textbf{VidaForge} \\
    \midrule[0.3pt]
    \rowcolor{wfprocessing}
    \multicolumn{5}{l}{\strut\textit{Video data processing}} \\
    \rowcolor{wfprocessingbody}
    Multi-node processing & \wfyes & \wfyes & \wfyes & \wfyes \\
    \rowcolor{wfprocessingbody}
    Scene clipping & \wfyes & \wfyes & \wfyes & \wfyes \\
    \rowcolor{wfprocessingbody}
    Quality filtering & \wfyes & \wfyes & \wfyes & \wfyes \\
    \rowcolor{wfprocessingbody}
    Perceptual near-duplicates & \wfno & \wfno & \wfno & \wfyes \\
    \rowcolor{wfprocessingbody}
    Semantic duplicates & \wfno & \wfyes & \wfyes & \wfyes \\
    \rowcolor{wfprocessingbody}
    Structured semantic tags & \wfyes & \wfno & \wfyes & \wfyes \\
    \addlinespace[4pt]
    \rowcolor{wfresearch}
    \multicolumn{5}{l}{\strut\textit{Data-recipe research}} \\
    \rowcolor{wfresearchbody}
    Reusable stage outputs & \wfyes & \wfyes & \wfyes & \wfyes \\
    \rowcolor{wfresearchbody}
    Accepted/rejected clip records & \wfyes & \wfyes & \wfyes & \wfyes \\
    \rowcolor{wfresearchbody}
    Full processing history & \wfno & \wfno & \wfno & \wfyes \\
    \rowcolor{wfresearchbody}
    Cross-model training & \wfno & \wfno & \wfno & \wfyes \\
    \rowcolor{wfresearchbody}
    Recipe $\rightarrow$ training $\rightarrow$ evaluation
      & \wfyes & \wfno & \wfno & \wfyes \\
    \bottomrule[0.6pt]
  \end{tabular}
  \endgroup
\end{table}

All four frameworks support reusable stage outputs and records for both
accepted and rejected clips, with different configuration requirements and
record contents. VidaForge organizes these capabilities around
recipe comparisons: it saves quality measurements separately from selection
decisions, retains rejected clips for alternative datasets, and links each
training sample to the outputs, parameters, and records of its processing
stages. The same processed pool connects to training readers for both
generative and self-supervised models.

The study in \cref{sec:evaluation} puts these capabilities into use:
saved scores and duplicate relations describe the differences among four
training recipes, retained rejected clips enable the Rejected comparison,
and the training paths carry the variants into Wan and V-JEPA~2 evaluation.
The sample examples in \cref{app:qualitative-examples} show the corresponding
processing decisions and annotations.

Coding agents may reduce the effort needed to add missing functionality, but
choosing which research workflows a framework should prioritize remains a
central design decision. VidaForge makes repeatable, traceable data-recipe
experiments a first-class requirement alongside processing throughput and
modality coverage.

\section{Conclusion}
\label{sec:conclusion}

VidaForge connects raw videos to traceable pretraining experiments, demonstrated
across generative and self-supervised learning and the million-clip
\textsc{VidaForge-3M} release.
Under fixed early-pretraining budgets, selection improves mean downstream
performance over random sampling at the same dataset size, while the full
Mixed pool achieves the highest mean scores on both benchmarks. The differing
loss and downstream rankings reinforce the value of model-level recipe evaluation.
Future work can study how scene splitting, filtering, deduplication, frame
sampling, and caption detail affect model performance, including with longer
training.
By sharing executable recipes and model-level evidence, academic groups can
build on one another's focused studies to develop a cumulative understanding
of video pretraining data. We hope VidaForge helps make this knowledge openly
accessible beyond the frontier labs with the resources to build such pipelines
independently.

\section*{Ethics Statement}

\textsc{VidaForge-3M} is derived from the publicly released
LLaVA-OneVision-2-Data~\citep{an2026llavaonevision2}, whose
\href{https://huggingface.co/datasets/mvp-lab/LLaVA-OneVision-2-Data}{dataset card}
specifies Apache-2.0. Our release uses the same license designation;
applicable upstream terms and third-party rights remain relevant to reuse.
VidaForge supports research on video pretraining data recipes. Users should
respect the privacy of individuals appearing in source videos and avoid
privacy-invasive or deceptive applications. Automatically generated captions
and labels may contain errors or biases and should not be treated as verified
facts about depicted individuals.

\section*{AI Use Statement}

We used generative AI tools to assist with writing, coding, and figure
preparation, and to provide feedback on the authors' experimental plans and
analyses. The models and procedures used for automatic dataset annotation
are described in \cref{sec:vidaforge,sec:evaluation}, with schemas and prompts
in \cref{app:annotation-schemas,app:annotation-prompts}.
The authors made the final research decisions and take responsibility for
the work.

\section*{Reproducibility Statement}

We release the VidaForge implementation and \textsc{VidaForge-3M} dataset
to support reproducible video data-recipe research.
\Cref{app:detailed-pipeline} details the processing stages and selection rules;
\cref{app:annotation-schemas,app:annotation-prompts} provide annotation schemas
and prompts. Dataset construction and validation isolation are described in
\cref{app:recipe_interventions,app:training-controls}, and training and
evaluation protocols and per-run downstream results are provided in
\cref{app:wan-details,app:vjepa-details}.
The implementation includes configurations and training adapters connecting
data recipes to Wan and V-JEPA~2 pretraining. We also provide a 2.5-hour
walkthrough video demonstrating how to use VidaForge.

\bibliography{iclr2027_conference}
\bibliographystyle{iclr2027_conference}

\FloatBarrier
\appendix
\section{Workflow Comparison Details}
\label{app:workflow-definitions}

The short labels in \cref{tab:framework-positioning} use the following
definitions. Multi-node processing, scene clipping, and quality filtering
refer to distributed video execution, scene-boundary clip construction,
and filtering with multiple quality signals, respectively.
\begin{itemize}
  \item \textbf{Reusable stage outputs:} processing results are saved and
  can be read by subsequent steps or runs. This does not require every
  internal operator to save a separate output or automatic dependency tracking.
  \item \textbf{Accepted/rejected clip records:} both sides of clip filtering
  can retain sample records and already-computed scores. This does not require
  scores from filters skipped after rejection or a complete processing history.
  Data-Juicer uses one-clip-per-record filtering with full tracing and score
  retention enabled; NeMo and Cosmos provide passed/filtered metadata writers.
  VidaForge writes all Select records alongside accepted and rejected subsets.
  \item \textbf{Perceptual near-duplicates:} detecting visually similar,
  non-identical clips. Data-Juicer's MD5 video-stream deduplication detects
  exact duplicates and does not satisfy this definition.
  \item \textbf{Semantic duplicates:} identifying related clips through
  semantic video representations.
  \item \textbf{Structured semantic tags:} video labels with defined fields
  or categories; free-form captions alone do not qualify.
  \item \textbf{Full processing history:} outputs, recipe parameters, and
  sample-level records saved at each stage and linked continuously to
  exported training data.
  \item \textbf{Cross-model training:} the same processed video pool
  connects to actual training readers for generative and self-supervised
  model families; multiple export formats alone do not qualify.
  \item \textbf{Recipe--training--evaluation:} an explicitly connected
  workflow constructs dataset variants, trains models on them, and
  evaluates the resulting models.
\end{itemize}

\paragraph{Comparison scope.}
The comparison follows a static review of code, configurations, documentation, and reported results from recent repository versions.  The Data-Juicer comparison includes Data-Juicer Sandbox~\citep{chen2025datajuicersandbox}, its extension for connecting data recipes to model training and evaluation.  This includes its integration with EasyAnimate~\citep{xu2024easyanimate}, a video-generation framework, and the supplied adaptations to EasyAnimate's training-data reader.  A dash can include partial support: for example, configuration backups, checkpoints, tracing, and stage replay are present in existing systems, while the stage-record row requires a continuous record through to training data.  VidaForge satisfies this definition through retained stage outputs and summaries.  The audit checks implemented connections; it does not rerun training or measure comparative speed.

\paragraph{Example: Studying a selection recipe.}
Consider the task in \cref{sec:evaluation}: construct different training
datasets from one video pool, compare their effects on video generation and
self-supervised representation learning, and retain each sample's processing
history. Our experiments use Wan and V-JEPA~2; the comparison concerns the two
types of learning. The reviewed public workflows provide different parts of
this task:
\begin{itemize}
  \item \textbf{Data-Juicer + Sandbox} connects recipe changes to video-generation
  training and evaluation. Completing this task requires adding a
  self-supervised training and evaluation path and linking the full processing
  history to the training samples.
  \item \textbf{NeMo Curator} provides video processing and dataset export.
  Completing this task requires connecting the dataset variants to model
  training and evaluation for both types of learning.
  \item \textbf{Cosmos Curator} provides video processing, reuse of saved
  processing results, and a video-generation training interface. Completing
  this task requires adding self-supervised training and connecting the
  recipe variants to model evaluation.
  \item \textbf{VidaForge} provides the connected dataset construction,
  training, and evaluation paths used in this study, together with the
  samples' processing records. Subsequent recipe comparisons can reuse
  these paths with new selection rules and training settings.
\end{itemize}

\FloatBarrier
\section{Detailed VidaForge Pipeline}
\label{app:detailed-pipeline}

This appendix follows the five-stage chain in \cref{eq:recipe-chain} and describes the steps that implement each transition.  Every step consumes a versioned directory of Parquet rows and writes another directory with the same sample identities, the inherited upstream record, and the fields owned by the current operation.  Steps that create media or model inputs also write the corresponding assets.  \texttt{video\_id} is assigned when a source video is first discovered, and \texttt{clip\_id} is assigned when a temporal interval is materialized as a clip.  Both identities are retained in all subsequent clip-level and training-level records.

Each output row distinguishes successful processing from a data-admission decision.  For example, \texttt{select\_ok} records whether the selection policy was evaluated, whereas \texttt{select\_pass} records whether the clip enters the selected view.  A failed operation still emits a row with its identity and an explicit error.  Each execution also records its \texttt{run\_id}, the \texttt{input\_run\_id} that it consumed, the step's recipe parameters, and aggregate counts in \texttt{summary.json}.  Following these links across the saved step outputs reconstructs the processing path of a training sample.  When execution resumes, completed rows are carried into the new output; asset-producing steps additionally verify that the expected file exists and is nonempty before reusing it.

\Cref{fig:recipe-flow} shows how many videos and clips continue or stop at
each processing step.  For an individual training clip, matching its
\texttt{clip\_id} and source \texttt{video\_id} across the saved outputs
recovers its source video, segmentation interval, selection measurements and
decision, annotations, and final packaged training entry.

\subsection{Stage I: Ingestion}
\label{app:stage-ingestion}

Ingestion transforms a heterogeneous raw-video snapshot $\mathcal{D}_0$ into a collection of standardized videos $\mathcal{D}_1$.  Its three steps separate media inspection, coarse eligibility rules, and media conversion, so each decision can be changed without rediscovering the raw collection.

\paragraph{Probe.}
Probe scans ordinary video files and videos stored as members of tar shards.  It creates one video row per discovered item, assigns a stable \texttt{video\_id} from its source identity and relative raw path, and records container, codec, duration, resolution, frame rate, bitrate, file size, and audio availability using \texttt{ffprobe}.  A tar member is identified by both the shard path and its path inside the shard.  Raw paths remain relative to the configured raw-data root, allowing the same metadata snapshot to be mounted at a different filesystem location.  Videos are inspected independently by Ray tasks~\citep{moritz2018ray}.  Probe failures retain their identities and error messages, giving Screen an explicit media-validity signal.

\paragraph{Screen.}
Screen evaluates configurable video-level rules over the Probe record.  The released recipe checks probe success, short-side resolution, frame rate, and duration; changing these YAML rules defines a different coarse source policy.  Each rule records the inspected value, predicate, and rejection reason in \texttt{screen\_json}.  Screen writes the complete evaluated population together with derived \texttt{pass/} and \texttt{reject/} views.  The distinction between \texttt{screen\_ok} and \texttt{screen\_pass} keeps execution failures separate from intentional rejection.  A new screening policy reuses Probe and reruns Screen, after which only the resulting input view needs to continue through the later stages.

\paragraph{Transcode.}
Transcode reads an explicit Screen view and creates the canonical video assets consumed by Segmentation.  It converts each input according to a configurable standardization profile covering video codec, frame rate, resolution, pixel format, and audio encoding.  The output is immediately probed again, so the row describes the finalized asset instead of inheriting codec or geometry fields from the source container.  Each video is converted by an independent Ray task and written under a path derived from \texttt{video\_id}.  Changes to codec, frame rate, or spatial normalization reuse Probe and Screen, then regenerate Transcode and the downstream states that consume the standardized video.

\subsection{Stage II: Segmentation}
\label{app:stage-segmentation}

Segmentation transforms each standardized video into scene-level clips $\mathcal{D}_2$.  Boundary estimation remains separate from clip creation, making the segmentation strategy inspectable before it changes the media assets and clip identities.

\paragraph{Detect.}
Detect analyzes each standardized video and records candidate scene boundaries as a list of timestamps.  The configured detector set can contain PySceneDetect variants~\citep{pyscenedetect}, TransNetV2~\citep{soucek2020transnetv2}, or uniform and seeded-random baselines.  When several frame-based detectors are selected, each processes the same video stream and their reported boundaries are combined.  The output records the detector names and resulting timestamps, while the full detector configuration is stored once in the run summary.  Detect creates the boundary record without generating new video files, and different videos are analyzed in parallel.  Consequently, multiple segmentation strategies can start from the same $\mathcal{D}_1$ and be inspected before Clip creates the corresponding video assets.

\paragraph{Clip.}
Clip converts the intervals defined by Detect into bounded-duration MP4 assets.  It subdivides overlong intervals, optionally trims context near genuine detector boundaries, and re-encodes each interval with FFmpeg to obtain accurate temporal cuts.  Every output receives a stable \texttt{clip\_id} and retains its parent \texttt{video\_id}, start and end times, detector record, and media properties.  Different source videos are processed concurrently by Ray tasks.  A changed Detect result changes the temporal units and therefore the clip identities; all clip-level stages that depend on those units are then recomputed, while the standardized-video state remains reusable.

\subsection{Stage III: Selection}
\label{app:stage-selection}

Selection transforms the complete clip population into a curated view $\mathcal{D}_3$.  VidaForge separates reusable context, measured evidence, duplicate relations, and the final admission policy.  This separation allows many threshold and quota variants to reuse the expensive measurements.

\paragraph{Context.}
Context prepares shared media inputs for downstream steps. The recipe uniformly samples frames at 2 fps and resizes them to a short side of 256 pixels while preserving the aspect ratio. It also extracts mono audio when the upstream media record indicates that an audio stream is present. Frames are required; audio remains optional. A single FFmpeg process produces both assets, which are stored in hash-bucketed directories keyed by \texttt{clip\_id}; their relative paths are recorded with the clip. The saved frames are reused by frame-based filtering, deduplication, and all three annotation steps---Camera, Caption, and Tag---without repeating frame extraction. Extracted audio is available to downstream steps configured to consume it. A change to frame sampling requires only the steps that consume these assets to be recomputed; the original clip remains unchanged.

\paragraph{Filter.}
Filter computes four independent clip-level measurements before Select decides dataset membership:
\begin{enumerate}
    \item \textbf{Optical quality} measures exposure and contrast over sampled frames and aggregates them conservatively across the clip.
    \item \textbf{Motion} reads the clip directly and combines FFmpeg VMAFMOTION~\citep{netflixvmaf} with scene-change detection, separating continuous motion from frequent shot transitions.
    \item \textbf{Aesthetics} applies Aesthetic Predictor V2.5~\citep{aestheticpredictorv25} to sampled frames and summarizes the lower-quality portion of the clip.
    \item \textbf{Visible text} uses a detection-only PP-OCRv5 model~\citep{cui2026ppocrv5} to estimate the image area occupied by text.  The score reverses this area fraction, so higher scores indicate less visible text.  It does not perform text recognition.
\end{enumerate}
All four scores lie in $[0,1]$, with higher values preferred by the selection recipe.
The following calculations describe the implementation and its default scoring parameters;
the study's admission thresholds are specified under Select below.
Let $G(x;a,b)=\min(1,\max(0,(x-a)/(b-a)))$ denote clipped linear growth,
$L(x;a,b)=1-G(x;a,b)$ linear decay, and $Q_p$ the $p$th percentile over sampled frames.

\textbf{Optical score.}
Each frame is converted to grayscale and scaled to $[0,1]$.
Its exposure score is $e=T(m)L(d;0.05,0.50)L(b;0.05,0.50)$,
where $m$ is median intensity, $d$ and $b$ are the fractions of pixels below
0.03 and above 0.97, respectively. The trapezoidal brightness function $T$
is zero outside $[0.05,0.95]$, one on $[0.20,0.80]$, and linear on the two ramps.
The frame contrast score is $c=G(I_{95}-I_5;0.05,0.25)$,
where $I_5$ and $I_{95}$ are pixel-intensity percentiles.
The clip score is $s_{\mathrm{optical}}=Q_{20}(e)Q_{20}(c)$:
exposure and contrast are aggregated separately before multiplication.

\textbf{Motion score.}
Let $v$ be mean VMAFMOTION over decoded clip frames and $r$ the fraction
of scene-detection frames whose FFmpeg \texttt{scdet} score reaches 10.
Scene detection uses 8 fps and width 320 with aspect ratio preserved.
The clip score is
$s_{\mathrm{motion}}=(1-2^{-\max(v,0)/8})L(r;0,0.05)$.
The first factor increases with motion strength, reaching 0.5 at $v=8$;
the second penalizes frequent scene transitions.

\textbf{Aesthetic and visible-text scores.}
Aesthetic Predictor outputs are mapped from $[1,10]$ to $[0,1]$ using $G$,
then aggregated with $Q_{20}$.
For text, detected regions with confidence at least 0.5 are rasterized into
a union mask so overlapping regions are counted once.
If $a$ is the fraction of image pixels covered by this mask,
$s_{\mathrm{text}}=1-Q_{95}(a)$. A score of 1 therefore indicates zero
text area at this clip-level percentile.

Each enabled filter records its processing status, normalized score, and detailed measurement evidence.  GPU-backed filters run in long-lived Ray workers that load each model once and process successive batches.  Filter choice and model configuration determine which measurements must be recomputed.  A change that only alters an admission threshold can reuse all Filter outputs and proceed directly to Select.

\paragraph{Dedup.}
Dedup computes two complementary duplicate relations:
\begin{enumerate}
    \item \textbf{PDQ}~\citep{meta2019pdq} computes 256-bit perceptual hashes for sampled frames and uses FAISS~\citep{douze2024faiss} exact binary Hamming search over the top 50 candidates per hash.  The study links a clip pair when at least 80\% of the sampled frames match within Hamming distance 31.
    \item \textbf{Cosmos-Embed}~\citep{nvidia2025cosmosembed1} computes one L2-normalized semantic embedding per clip from uniformly sampled frames.  FAISS with a cuVS backend~\citep{nvidia2026cuvs} retrieves the top 50 candidates by inner product; the study links pairs with cosine similarity at least 0.95.
\end{enumerate}
For each method, matched clip pairs define edges in a graph whose nodes are clips. Union-find merges connected clips into candidate duplicate groups. Clips can belong to the same group through intermediate matches, even when they do not directly match each other. PDQ and Cosmos-Embed retain separate group identities, keeping their visual and semantic evidence independently inspectable.

Dedup executes feature extraction and global matching as two saved phases.  Multiple feature-extraction workers compute hashes or embeddings in parallel and write feature shards.  Lightweight clip information is stored in Parquet, while hash and embedding matrices are stored separately as arrays.  Each matching worker then loads the complete feature store and builds the corresponding search index.  The clip population is partitioned into non-overlapping query ranges, which the workers search in parallel.  Replicating the index simplifies execution: each worker retrieves global top-$k$ candidates independently, avoiding cross-shard query routing, communication, and candidate merging.  This design trades per-worker index memory for parallel query processing; adding workers divides the query workload while retaining the full index on each worker.  At larger scales, sharded indexes or a distributed vector search service could reduce per-worker memory requirements.  The matched pairs are combined to form the groups described above.  Each clip receives a group identifier, group size, and matching evidence; the high-dimensional features remain in the feature store.  Feature extraction can be skipped when an existing store is available, so a new matching threshold or index configuration can reuse the saved features.  Sample admission remains the responsibility of Select.

\paragraph{Select.}
Select combines the Filter measurements and Dedup relations through one YAML policy.  Score rules define clip-level quality gates, while per-group quotas determine how many members of each duplicate group remain eligible.  For every clip, Select records whether it is admitted, the first rule that rejects it when applicable, and the outcome of every evaluated rule.  Select preserves the complete input population and additionally writes \texttt{pass/} and \texttt{reject/} views, so the measured properties of both populations remain available for inspection and alternative sampling.  Changing a score threshold, enabling a previously computed signal, or adjusting a duplicate-group quota requires only a new Select execution.  Later annotation and packaging steps can then consume the corresponding view without repeating Context, Filter, or Dedup.

The coverage--quality study requires optical, motion, aesthetic, and visible-text scores of at least 0.9, 0.1, 0.1, and 0.5.
Among clips passing these gates, both deduplication rules rank candidates by
the unweighted sum of the four scores, in descending order, breaking ties
by ascending lexicographic order of \texttt{clip\_id}.
PDQ retains the highest-ranked eligible clip per group.
Among clips that pass quality filtering and PDQ selection, we rank members
of each Cosmos-Embed candidate group using the same ordering.
The initial retained set contains 20\% of these candidates, rounded up and bounded
between 1 and 20 clips. We compare every remaining candidate with this fixed
initial set using the saved Cosmos-Embed embeddings. We exclude a candidate only
when its maximum cosine similarity to the initial set is at least 0.95.
We retain candidates without a qualifying match, so the final retained count
may exceed the initial quota. We keep the comparison set fixed throughout this
check; additionally retained clips do not become references.

\subsection{Stage IV: Annotation}
\label{app:stage-annotation}

Annotation augments the clips chosen for a dataset variant with descriptions and structured labels, producing $\mathcal{D}_4$.  The variant is defined by the upstream selection records, and all three steps retain the quality measurements, duplicate relations, selection decisions, and sample identities.  Camera, Caption, and Tag issue bounded concurrent requests to Ray-managed OpenAI-compatible vLLM servers~\citep{kwon2023vllm} and record their model, prompt, and schema versions together with processing and validation outcomes.  \Cref{app:annotation-schemas,app:annotation-prompts} provide the current output schemas and prompts.

\paragraph{Camera.}
Camera adapts the CameraBench camera-motion taxonomy~\citep{lin2025camerabench} to describe intentional motion, steadiness, rotation, translation, zoom, object-centric tracking, motion speed, cinematic effects, and scene dynamics.  The vision-language model predicts labels from the corresponding category sets, and a response that violates this contract is recorded as an annotation failure.  Changing the camera prompt, schema, or model starts from the saved Context and selection records.

\paragraph{Caption.}
Caption produces four progressively detailed natural-language descriptions in one vision-language-model request. It directly observes the saved Context frames and receives the Camera record as auxiliary structured context. The prompt instructs the model to describe observable content and express useful camera information naturally, without copying raw labels. The implementation supports optional audio input, but the experiments in this paper do not supply audio to Caption. The four caption levels remain separate fields, allowing target training code to select a desired level or apply its own caption policy. The complete model response is preserved. A caption variant reuses the prepared media context and upstream camera output, then reruns only the annotation and packaging operations that consume the changed text.

\paragraph{Tag.}
Tag adds structured fields for domain, scene, subjects, actions, visual style, visible text, and watermark presence.  Its prompt reads the sampled frames and clip duration directly.  Caption and Camera outputs are inherited in the row but are not included in the Tag prompt, preventing their wording from determining the category labels.  Subjects and actions support multiple labels, while the remaining fields use a single label.  This dependency structure allows Tag to change without regenerating Camera or Caption outputs.

\subsection{Stage V: Packaging}
\label{app:stage-packaging}

Packaging transforms annotated clips into the target-specific training dataset $\mathcal{D}_5$.  It consumes the explicit selection and annotation fields produced upstream and implements the data contract expected by a training repository.  We instantiate this stage with two representative paths for video foundation models: video generation with Wan~2.1~\citep{wanteam2025wan} and self-supervised representation learning with V-JEPA~2~\citep{assran2025vjepa2}.  Both are derived from the same $\mathcal{D}_4$, allowing upstream media processing, selection, and annotation results to be reused across training systems.

\paragraph{Wan through NeMo-AutoModel.}
The Wan~2.1 path pairs each clip with its caption and converts them into the cached representations expected by NeMo-AutoModel~\citep{nvidia2026nemoautomodel}.  Clip duration, frame rate, resolution, and aspect ratio determine its temporal and spatial bucket.  GPU workers then compute video latents with the Wan VAE and text embeddings with the text encoder.  Each cached record stores these representations, its bucket assignment, and the video and clip identities needed to trace it back to the upstream recipe.  A lightweight index allows the training loader to locate the caches and batch samples with compatible shapes.

\paragraph{V-JEPA~2.}
The V-JEPA~2 path illustrates a different training-data contract: its self-supervised objective consumes video clips without captions or precomputed representations.  VidaForge retains clips that meet the model-specific duration and resolution requirements, then writes the video manifest used by the released V-JEPA~2 training code.  A TorchCodec-based dataset~\citep{metapytorch2025torchcodec} reads this manifest during training.  The packaged records preserve each clip's identity, selection evidence, segmentation interval, and source video identity.

\FloatBarrier
\section{Annotation Schemas}
\label{app:annotation-schemas}

Stage IV uses explicit schemas to define valid outputs and acceptance checks.  Camera, Caption, and Tag execute in sequence: Caption consumes the Camera result, while Tag classifies the sampled frames without using the wording produced by the preceding steps.  Every annotation records its model, schema, and prompt versions; parsing and validation failures remain explicit under the same clip identity.  \Cref{app:annotation-prompts} reproduces the corresponding prompts.

\subsection{Camera-Motion Taxonomy}
\label{app:camera-schema}

Camera receives ordered sampled frames with their timestamps, clip duration, and sampling specification.  The Camera schema adapts the CameraBench taxonomy of camera-, ground-, and object-centric motion primitives~\citep{lin2025camerabench} into the fields in \cref{tab:camera-schema}.  VidaForge adds \texttt{unknown} for insufficient visual evidence.  Cinematic effects are multi-label; all other leaf fields are single-label.

\FloatBarrier
\begin{table}[ht!]
    \centering
    \small
    \caption{\textbf{Camera schema.} Nested groups contain the listed leaf fields.  Every leaf field also permits \texttt{unknown}.}
    \label{tab:camera-schema}
    \begin{tabular}{p{0.15\linewidth} p{0.22\linewidth} p{0.53\linewidth}}
        \toprule
        Group & Fields & Allowed labels \\
        \midrule
        Motion type & \texttt{motion\_type} & \texttt{no-motion}, \texttt{minor-motion}, \texttt{simple-motion}, \texttt{complex-motion} \\
        Steadiness & \texttt{steadiness} & \texttt{static}, \texttt{no-shaking}, \texttt{minimal-shaking}, \texttt{unsteady}, \texttt{very-unsteady} \\
        Rotation & \texttt{pan}, \texttt{tilt}, \texttt{roll} & \texttt{pan-left/right/no-pan}; \texttt{tilt-up/down/no-tilt}; \texttt{roll-CW/CCW/no-roll} \\
        Translation & \texttt{dolly}, \texttt{pedestal}, \texttt{truck} & \texttt{dolly-in/out/no-dolly}; \texttt{pedestal-up/down/no-pedestal}; \texttt{truck-left/right/no-truck} \\
        Intrinsic & \texttt{zoom} & \texttt{zoom-in}, \texttt{zoom-out}, \texttt{no-zoom} \\
        \multirow[t]{3}{*}{Object-centric} & \texttt{arc} & \texttt{arc-CW}, \texttt{arc-CCW}, \texttt{no-arc} \\
        & \texttt{arc\_tracking}, \texttt{lead\_tracking}, \texttt{tail\_tracking}, \texttt{side\_tracking}, \texttt{aerial\_tracking}, \texttt{pan\_tracking}, \texttt{tilt\_tracking} & \texttt{arc-tracking/no-arc-tracking}; \texttt{lead-tracking/no-lead-tracking}; \texttt{tail-tracking/no-tail-tracking}; \texttt{side-tracking/no-side-tracking}; \texttt{aerial-tracking/no-aerial-tracking}; \texttt{pan-tracking/no-pan-tracking}; \texttt{tilt-tracking/no-tilt-tracking} \\
        & \texttt{subject\_size\_change} & \texttt{subject-larger}, \texttt{subject-smaller}, \texttt{no-subject-change} \\
        Speed & \texttt{speed} & \texttt{slow}, \texttt{regular}, \texttt{fast}, \texttt{none} \\
        Effects & \texttt{effects} & Any observed subset of \texttt{frame-freezing}, \texttt{dolly-zoom}, and \texttt{motion-blur}; otherwise only \texttt{none} or only \texttt{unknown} \\
        Scene dynamics & \texttt{scene\_dynamics} & \texttt{static}, \texttt{mostly-static}, \texttt{dynamic} \\
        \bottomrule
    \end{tabular}
\end{table}
\FloatBarrier

The response must match the nested JSON schema exactly.  Extra fields, invalid labels, and incompatible effect combinations fail validation.

\subsection{Multi-Level Caption Specification}
\label{app:caption-schema}

Caption receives the ordered frames, timestamps, clip duration, structured Camera result, and, in \texttt{video\_audio} mode, extracted audio.  A single request returns the four strings in \cref{tab:caption-schema}.  Their separate fields let a packaging recipe choose the amount of text supplied to a target model.

\FloatBarrier
\begin{table}[ht!]
    \centering
    \small
    \caption{\textbf{Caption schema.} Target lengths summarize the intended granularity of the four caption levels, informed by observed caption lengths.}
    \label{tab:caption-schema}
    \begin{tabular}{p{0.11\linewidth} p{0.15\linewidth} p{0.64\linewidth}}
        \toprule
        Field & Target length & Required content \\
        \midrule
        \texttt{level\_0} & $<20$ & Main subject, primary action, and scene; camera motion and audio are omitted. \\
        \texttt{level\_1} & 20--50 words & Concise event flow, important subjects and changes, noticeable camera motion, and a summary of important audio when available. \\
        \texttt{level\_2} & 50--100 words & Initial state, temporal progression, subject--object relations, position changes, background, main camera motion, and summarized speech or other relevant audio. \\
        \texttt{level\_3} & 100--200 words & Dense reconstruction of subjects, attributes, actions, environment, lighting, style, composition, temporal changes, camera behavior, visible text, and relevant audio.  In \texttt{video\_audio} mode, clearly audible speech is included. \\
        \bottomrule
    \end{tabular}
\end{table}
\FloatBarrier

The Caption schema accepts exactly these four strings.  The saved annotation also identifies its mode, model, schema version, and prompt version.

\subsection{Semantic Tag Taxonomy}
\label{app:tag-schema}

Tag receives the sampled frames, timestamps, and clip duration without Camera or Caption text.  Its category decisions therefore form a separate annotation view over the visual evidence.  In the Tag schema in \cref{tab:tag-schema}, subjects and actions are multi-label and the remaining fields use one dominant label.

\FloatBarrier
\begin{table}[ht!]
    \centering
    \small
    \caption{\textbf{Tag schema.} \texttt{subjects} and \texttt{actions} are multi-label fields; all other fields are single-label.}
    \label{tab:tag-schema}
    \begin{tabular}{p{0.14\linewidth} p{0.78\linewidth}}
        \toprule
        Field & Allowed labels \\
        \midrule
        \texttt{domain} & \texttt{real\_world}, \texttt{animation}, \texttt{game}, \texttt{screen\_recording}, \texttt{synthetic\_render}, \texttt{mixed}, \texttt{unknown} \\
        \texttt{scene} & \texttt{general\_indoor}, \texttt{general\_outdoor}, \texttt{urban}, \texttt{nature}, \texttt{driving}, \texttt{sports}, \texttt{food}, \texttt{product}, \texttt{portrait}, \texttt{screen}, \texttt{other}, \texttt{unknown} \\
        \texttt{subjects} & \texttt{person}, \texttt{vehicle}, \texttt{animal}, \texttt{object}, \texttt{food}, \texttt{landscape}, \texttt{building}, \texttt{text}, \texttt{screen}, \texttt{robot}, \texttt{other}, \texttt{unknown} \\
        \texttt{actions} & \texttt{talking}, \texttt{locomotion}, \texttt{driving}, \texttt{sports}, \texttt{cooking}, \texttt{object\_manipulation}, \texttt{natural\_motion}, \texttt{camera\_motion\_only}, \texttt{timelapse}, \texttt{none}, \texttt{other}, \texttt{unknown} \\
        \texttt{style} & \texttt{photorealistic}, \texttt{cinematic}, \texttt{documentary}, \texttt{anime}, \texttt{cartoon}, \texttt{cg\_render}, \texttt{gameplay}, \texttt{graphic}, \texttt{unknown} \\
        \texttt{text} & \texttt{none}, \texttt{incidental}, \texttt{subtitle}, \texttt{screen\_ui}, \texttt{document}, \texttt{signage}, \texttt{overlay\_text}, \texttt{unknown} \\
        \texttt{watermark} & \texttt{none}, \texttt{logo}, \texttt{text\_watermark}, \texttt{platform\_watermark}, \texttt{unknown} \\
        \bottomrule
    \end{tabular}
\end{table}
\FloatBarrier

Validation prevents \texttt{unknown} from accompanying another label and \texttt{none} from accompanying another action.  The validated JSON object is stored together with separate tag fields for distribution analysis and sampling.

\FloatBarrier
\section{Annotation Prompts}
\label{app:annotation-prompts}

This appendix presents the versioned Stage IV prompts, with a stationary-camera speed label in \cref{app:camera-prompt} and revised caption-length targets in \cref{app:caption-prompt}.  Angle-bracketed tokens denote values supplied for an individual clip.  \texttt{<FRAME\_TIMELINE>} expands to one ordered \texttt{frame\_XXXX: timestamp} line per sampled frame; \texttt{<CAMERA\_CONTEXT>} expands to the validated Camera fields; and \texttt{<ORDERED\_IMAGE\_INPUTS>} and \texttt{<AUDIO\_INPUTS>} denote the multimodal inputs attached to the user message.  Tokens ending in \texttt{\_JSON\_SCHEMA>} refer to the corresponding strict output contracts in \cref{app:annotation-schemas}; the schema is not duplicated inside the listings.

\subsection{Camera QA}
\label{app:camera-prompt}

\begin{lstlisting}[style=annotationprompt]
[SYSTEM MESSAGE]
You are a camera-motion annotation model.
You must infer camera movement from the provided ordered frame sequence.
Return only valid JSON. Do not include markdown, comments, explanations, or confidence scores.

[USER TEXT MESSAGE]
Task: annotate camera motion for this clip from the ordered frames.

You will receive <NUM_FRAMES> image frames in chronological order.
Clip duration: <DURATION_SEC> seconds.
Frame sampling: <SAMPLED_FPS> fps, method=<SAMPLING_METHOD>.

Frame timestamps:
<FRAME_TIMELINE>

Important rules:
- Judge camera motion, not subject motion.
- Keep intentional camera movement separate from unintended shake.
- If the visual evidence is insufficient for a dimension, output "unknown".
- Use "no-*" labels only when the motion is explicitly absent.
- Do not output confidence scores.
- Return exactly one JSON object matching the schema below.

Label definitions and guidelines:

Motion type:
- no-motion: The camera remains stationary with no intentional movement. Note: Unintentional shaking belongs to no-motion.
- minor-motion: The camera moves slightly and intentionally, such as a gentle pan or zoom. The motion is noticeable but remains subtle and not significant.
- simple-motion: The camera moves significantly in a straightforward manner, such as a steady pan, tilt, arc, or simple tracking shot. Select this even if the video combines two or more motions, as long as they occur simultaneously at roughly the same speed.
- complex-motion: The camera exhibits complex movements that are difficult to classify. This includes conflicting motion, sequential motion, simultaneous motions at different speeds, or unclear motion / missing background information due to motion blur or lack of background cues.

Steadiness:
- static: The camera remains completely stationary with no movement or vibration.
- no-shaking: The camera moves smoothly with no detectable shake, typically using high-end stabilizers. Select only if the camera is moving and no unintended motion is present.
- minimal-shaking: The camera exhibits slight shaking, whether stationary or moving, maintaining a mostly stable shot.
- unsteady: The camera shows moderate shaking, whether stationary or in motion, introducing noticeable but controlled instability.
- very-unsteady: The camera shakes consistently, typical of unstabilized handheld or action footage. Select only if shaking is consistent throughout the video.

Translation:
- dolly-in / dolly-out: The camera moves forward or backward relative to the ground plane and the initial frame.
- no-dolly: The camera does not move forward/backward during the shot.
- pedestal-up / pedestal-down: Select this when the camera moves upward or downward clearly and consistently relative to the ground or the orientation of the initial frame.
- no-pedestal: The camera does not move upward or downward during the shot.
- truck-left / truck-right: The camera physically moves to the left or right, changing its position relative to the initial frame.
- no-truck: The camera does not move to the left or right during the shot.

Rotation:
- pan-left / pan-right: The camera rotates its angle by pivoting left or right with respect to the initial frame.
- no-pan: The camera does not pan left or right.
- tilt-up / tilt-down: The camera rotates its angle up or down vertically with respect to the initial frame.
- no-tilt: The camera does not tilt up or down.
- roll-CW / roll-CCW: The camera performs a clear and consistent clockwise (CW) or counterclockwise (CCW) roll by rotating around its own optical center.
- no-roll: The camera does not roll clockwise/counterclockwise.

Intrinsic change:
- zoom-in / zoom-out: The camera adjusts its focal length to zoom in or out, changing the frame size. This differs from physical camera movement.
- no-zoom: The camera does not adjust its focal length during the video.

Object-centric movement:
- arc-CW / arc-CCW: The camera moves in a circular or semi-circular motion around the subject or the frame center in a clockwise or counterclockwise direction.
- no-arc: The camera does not move in a circular or semi-circular motion during the video.
- arc-tracking: The camera moves in a circular or semi-circular path around the moving subject, often referred to as an orbit or circular tracking shot.
- no-arc-tracking: The camera does not track or does not move in a circular or semi-circular path around the moving subject.
- lead-tracking: The camera moves ahead of the moving subject, capturing their face or front as they follow the camera's path. This is also referred to as a leading shot.
- no-lead-tracking: The camera does not track or does not move ahead of the moving subject.
- tail-tracking: The camera follows directly behind the moving subject, keeping their back in view as they move forward. This is also known as a follow shot or chase shot.
- no-tail-tracking: The camera does not track or does not move behind the moving subject.
- side-tracking: The camera moves parallel to the moving subject, following them from the side as they move through the scene.
- no-side-tracking: The camera does not track or does not move parallel to the moving subject.
- aerial-tracking: The camera tracks the moving subject from a high vantage point, often using a drone or crane to follow their movement.
- no-aerial-tracking: The camera either does not track the moving subject or is not positioned at a high vantage point.
- pan-tracking: The camera remains in a fixed position but pivots horizontally to follow the subject as they move.
- no-pan-tracking: The camera does not track the subject or does not pivot horizontally to follow their movement.
- tilt-tracking: The camera tilts up or down to follow the vertical movement of the subject.
- no-tilt-tracking: The camera does not track the subject or does not pivot vertically to follow their movement.
- subject-larger: The camera moves or zooms in towards the tracked subject, making them appear larger in the frame.
- subject-smaller: The camera moves or zooms away from the tracked subject, making them appear smaller in the frame.
- no-subject-change: The camera neither moves towards nor away from the subject.

Camera movement speed:
- slow: The camera moves at a noticeably slow pace.
- regular: The camera moves at a regular pace. If the speed does not stand out as particularly slow or fast, it is considered regular.
- fast: The camera moves quickly, such as in a crash zoom or whip pan.
- none: No camera movement is visible; the camera remains stationary.

Cinematic motion effects:
- frame-freezing: A visual effect where scene motion is paused or frozen mid-action, creating a still frame within a moving sequence.
- dolly-zoom: A camera effect where the background appears to compress or stretch while the subject stays the same size, often used to create a sense of unease.
- motion-blur: A visual effect where moving objects blur due to slow shutter speed or camera movement, often used to emphasize speed and fluid motion in action scenes.

Scene dynamics:
- static: The entire scene, including all subjects and background, remains completely motionless throughout the video.
- mostly-static: The scene is largely still, with only minor elements or small parts exhibiting movement.
- dynamic: A significant portion of the frame is occupied by dynamic movement of subjects or scene elements, excluding camera motion, that visibly alters the scene.

Schema-specific fallback labels:
- unknown: Use only when the visual evidence is insufficient for that field. This fallback is added by this project; it is not listed in the source table.
- effects none: Use only when no cinematic motion effect is visible. This fallback is added by this project; it is not listed in the source table.
- effects is a multi-label field in this project. Output every visible cinematic motion effect, or output only ["none"] / only ["unknown"]. Do not mix none or unknown with other effect labels.

Allowed labels and expected JSON fields:
<CAMERA_JSON_SCHEMA>

[USER MEDIA]
<ORDERED_IMAGE_INPUTS>

[STRICT OUTPUT CONTRACT]
<CAMERA_JSON_SCHEMA>
\end{lstlisting}

\subsection{Multi-Level Caption}
\label{app:caption-prompt}

The caption prompt has separate substitutions for video-only and video-audio operation.  Both substitutions are included below.

\begin{lstlisting}[style=annotationprompt]
[SYSTEM MESSAGE]
You are a video captioning model.
You must describe only observable evidence from the provided media.
Return only valid JSON. Do not include markdown, comments, or explanations.

[USER TEXT MESSAGE]
Task: generate four multi-level captions for this video clip.

Clip context:
Clip duration: <DURATION_SEC> seconds.

Frame context:
Image frames: <NUM_FRAMES> frames in chronological order.
Frame sampling: <SAMPLED_FPS> fps, method=<SAMPLING_METHOD>.
Frame timestamps:
<FRAME_TIMELINE>

Audio context:
<AUDIO_CONTEXT>

Camera context:
<CAMERA_CONTEXT>

Caption levels:
- level_0: Write a semantic gist in <20 words. Describe only the main subject, primary action, and scene. Omit camera motion and audio.
- level_1: Write a concise video caption in 20-50 words. Summarize the main event, key subjects and actions, setting, and major changes. Mention noticeable camera motion and important audio when present.
- level_2: Write a detailed temporal caption in 50-100 words. Describe the initial state and how the action unfolds, including subject-object relationships, position changes, relevant background, and main camera motion. Summarize speech and other relevant sounds when useful.
- level_3: Write a dense reconstruction caption in 100-200 words. Describe subjects, attributes, actions, setting, lighting, color, style, composition, spatial relationships, temporal changes, camera motion, visible watermarks or logos, and relevant audio. In video_audio mode, include clearly audible speech in full when present, introduced naturally, for example, 'He says: ...'.

General rules:
- Focus on observable content. Do not infer hidden intent, symbolism, or backstory.
- Preserve temporal order. Describe what changes from the beginning to the middle and end.
- Describe subject motion and camera motion separately when possible.
- Use camera context from level_1 onward, but write it naturally. Never copy raw camera labels into the caption.
- For level_1, mention camera motion only if it is visually noticeable and relevant.
- For level_2, integrate the main camera motion into the action progression.
- For level_3, use all useful camera context to describe cinematography, framing, and motion dynamics.
- Avoid over-emphasizing tiny shake or small camera adjustments unless they affect the clip.
- Mention watermark/logo/text only if visible.
- The four captions should be different levels of detail, not near-duplicates.
- Do not include explicit frame IDs or timestamp numbers in the captions.
- Return exactly one JSON object with keys in this order: level_3, level_2, level_1, level_0.
- Each value must be a single string.

[AUDIO_CONTEXT: VIDEO MODE OR NO AUDIO]
No audio is provided.

[AUDIO_CONTEXT: VIDEO_AUDIO MODE]
Use the provided audio as supporting evidence. If speech is present, describe who is speaking when visible, the speech context, and the semantic content of the speech. Level_3 must include the complete clearly audible speech content when speech is present, while still remaining a dense audio-visual caption rather than a standalone transcript. Put the spoken words in level_3 using a natural phrase such as 'He says: ...' or 'She says: ...'. Do not satisfy this requirement by only writing phrases such as 'he speaks', 'speech is clear', or 'speech is intelligible'. If speech is audible but you cannot reliably identify the words, write this exact sentence in level_3: 'The exact speech content is not clearly audible.' If speech is long, include as much as can be reliably heard without inventing missing words. For level_1 and level_2, summarize the speech content instead of transcribing it fully. Also mention music, sound effects, ambient sound, or silence when they are audible and relevant. If the audio is unclear, do not guess.

[CAMERA_CONTEXT: NO CAMERA RECORD]
No external camera context is provided.

[CAMERA_CONTEXT: CAMERA RECORD AVAILABLE]
<CAMERA_JSON_WITH_LABEL_FIELDS>

[USER MEDIA]
<ORDERED_IMAGE_INPUTS>
<AUDIO_INPUTS: VIDEO_AUDIO MODE ONLY>

[STRICT OUTPUT CONTRACT]
<CAPTION_JSON_SCHEMA>
\end{lstlisting}

\subsection{Semantic Tags}
\label{app:tag-prompt}

\begin{lstlisting}[style=annotationprompt]
[SYSTEM MESSAGE]
You are a video semantic tagging model.
You must assign stable low-cardinality tags from the provided ordered frame sequence.
Return only valid JSON. Do not include markdown, comments, explanations, or confidence scores.

[USER TEXT MESSAGE]
Task: assign clip-level semantic tags for this video clip.

Clip context:
Clip duration: <DURATION_SEC> seconds.

Frame context:
Image frames: <NUM_FRAMES> frames in chronological order.
Frame sampling: <SAMPLED_FPS> fps, method=<SAMPLING_METHOD>.
Frame timestamps:
<FRAME_TIMELINE>

Important rules:
- Assign tags for the whole clip, not for a single isolated frame.
- Do not output confidence scores.
- Return exactly one JSON object matching the schema below.

Label definitions and guidelines:

Domain:
- real_world: Camera-captured real-world footage, including phones, professional cameras, dashcams, drones, and surveillance-like footage. If a physical screen is filmed by a camera, use real_world and include screen in subjects.
- animation: Authored animated video, including 2D/3D animation, anime, cartoons, animated characters, and stop-motion-like animation.
- game: Video game content or game engine gameplay, including first-person, third-person, and game replay footage, even when captured from a screen.
- screen_recording: Direct digital capture of software UI, websites, documents, slides, terminals, phone/computer interfaces, or recorded presentations. If the dominant content is gameplay, use game instead. If a physical screen is filmed by a camera, use real_world instead.
- synthetic_render: Non-game CGI, 3D render, simulation, product render, or synthetic scene that is not authored animation or gameplay.
- mixed: Multiple domains are visually central or occupy a substantial portion of the clip, such as real footage with large animation overlays or split-screen mixed sources. Do not use mixed for small overlays, stickers, captions, subtitles, logos, or minor inserted graphics.
- unknown: The data form cannot be determined from the evidence.

Scene:
- Scene is single-label. Choose the dominant scene/content bucket. Use general_indoor/general_outdoor only when no more specific scene/content label dominates.
- general_indoor: Main scene is inside a building or enclosed space, and no more specific scene label dominates.
- general_outdoor: Main scene is outside, and no more specific scene label dominates.
- urban: City, street, road-side pedestrian, architecture, crowd, or built-environment scene where driving/navigation is not the main focus.
- nature: Landscape, forest, mountain, ocean, sky, plants, wildlife habitat, or natural environment.
- driving: Road, traffic, vehicle-mounted, dashcam, cockpit, or vehicle-centered driving/navigation scene where the road, vehicle, or navigation context is the main focus.
- sports: Organized sport, exercise, fitness, competition, or athletic activity scene.
- food: Food preparation, eating, restaurant, ingredients, or food-focused scene.
- product: Product/object showcase, commercial-like object shot, package, appliance, tool, or item demonstration.
- portrait: Person-centered close-up, selfie, talking-head, interview, face/body portrait, or presenter-dominant scene.
- screen: Screen/UI/document/slides dominate the visual content.
- other: A clear scene exists but does not fit the listed labels.
- unknown: The scene cannot be determined from the evidence.

Subjects:
- person: One or more visible humans or human-like characters.
- vehicle: Car, truck, bus, train, bicycle, motorcycle, aircraft, boat, or similar vehicle.
- animal: Any visible animal.
- object: A salient foreground physical object, tool, toy, product, device, furniture, or item. Do not add object for ordinary background clutter.
- food: Food, drink, ingredients, dishes, or cooking material.
- landscape: Natural scenery, sky, water, mountain, forest, plants, or terrain as a main subject.
- building: Buildings, rooms, streetscape structures, bridges, monuments, or architecture.
- text: Text itself is a salient visible subject, such as a sign, poster, title card, slide text, document text, or large overlay text.
- screen: A visible screen or screen-captured interface is a salient subject, such as a monitor, phone display, TV, software UI, webpage, document, slide, or game HUD.
- robot: Robot, robotic arm, autonomous machine, or embodied AI platform.
- other: A clear major subject exists but does not fit the listed labels.
- unknown: Use only when the visible subject cannot be determined.

Actions:
- talking: A person or character appears to speak, present, sing, or address the camera/audience.
- locomotion: Walking, running, hiking, dancing steps, jumping, or human/character movement through space.
- driving: Vehicle driving, riding, traffic movement, or road navigation.
- sports: Sport, exercise, fitness, competition, or athletic motion.
- cooking: Cooking, food preparation, plating, eating preparation, or kitchen action.
- object_manipulation: Hands, tools, robots, or subjects manipulate, assemble, open, move, or operate objects.
- natural_motion: Natural motion such as flowing water, fire, smoke, weather, plants moving, or animal/environment motion.
- camera_motion_only: Camera movement is the main visible change while subjects/scene are mostly static. Do not use this for minor camera shake when another action is visible.
- timelapse: Time-lapse, accelerated growth, sunset, traffic trails, construction progress, or clearly sped-up temporal change.
- none: No meaningful subject or scene action is visible.
- other: A clear action exists but does not fit the listed labels.
- unknown: The action cannot be determined from the evidence.

Style:
- photorealistic: Realistic-looking footage or synthetic imagery that is not clearly documentary, cinematic, gameplay, graphic, anime/cartoon, or cg_render.
- cinematic: Deliberately stylized camera, lighting, color grading, composition, or film-like visual treatment. Use only when the visual treatment clearly stands out.
- documentary: Factual, vlog, news, tutorial, documentation-like, or observational real-world footage, without strong cinematic stylization.
- anime: Anime-style visual content.
- cartoon: Cartoon, 2D animation, western animation, simple animated characters, or non-anime cartoon style.
- cg_render: CGI, 3D render, simulation, digital twin, product render, or synthetic 3D scene.
- gameplay: Game visual style, including HUD/game engine aesthetics.
- graphic: Slides, diagrams, charts, infographic, UI-heavy graphic design, or mostly flat graphical content.
- unknown: The visual style cannot be determined from the evidence.

Text:
- Text is single-label. Choose the dominant visible text role.
- none: No visible text.
- incidental: Small or background text that is not central, such as labels, packaging, signs in the distance, or minor UI text.
- subtitle: Subtitles, captions, karaoke lyrics, or dialogue text overlaid on video. Use this when subtitles/captions are persistent and central.
- screen_ui: Software UI, websites, apps, menus, code editors, terminals, game HUDs, interface controls, or phone/computer screen text.
- document: PPT/PDF/slides, paper pages, books, posters, forms, presentation pages, or page-like document content, even when viewed on a screen.
- signage: Signs, billboards, road signs, storefront signs, placards, or wayfinding text.
- overlay_text: Large title text, meme text, sticker text, callouts, or other prominent text overlaid on the video that is not subtitle, UI, document, or signage.
- unknown: Visible text role cannot be determined.

Watermark:
- none: No visible watermark or logo-like ownership/platform mark.
- logo: A visible overlay/source/channel logo mark. Do not use this for ordinary logos printed on physical products or signs.
- text_watermark: Repeated, semi-transparent, or ownership text watermark.
- platform_watermark: Platform/app/source watermark, such as social media or stock-media marks.
- unknown: Watermark/logo status cannot be determined.

General rules:
- Single-label fields: domain, scene, style, text, watermark.
- Multi-label fields: subjects, actions.
- Choose the dominant label for single-label fields.
- Dominant means sustained across the clip and visually/semantically central, not a brief or background appearance.
- Multi-label fields should include all major visible labels, but stay concise.
- Use other when the evidence is clear but does not fit the listed labels.
- Use unknown only when the evidence is insufficient.
- Do not mix unknown with any other label in multi-label fields.
- Do not mix none with any other action label.

Allowed labels and expected JSON fields:
<TAG_JSON_SCHEMA>

[USER MEDIA]
<ORDERED_IMAGE_INPUTS>

[STRICT OUTPUT CONTRACT]
<TAG_JSON_SCHEMA>
\end{lstlisting}

\FloatBarrier
\section{VidaForge-3M}
\label{app:vidaforge3m}

\subsection{Dataset Construction and Overview}
\label{app:vidaforge3m-overview}

\textsc{VidaForge-3M} is constructed from the video portion of LLaVA-OneVision-2-Data~\citep{an2026llavaonevision2}.\footnote{\url{https://huggingface.co/datasets/mvp-lab/LLaVA-OneVision-2-Data}}  We process 800,000 source videos through the five-stage VidaForge pipeline and release 3,141,246 scene-level clips totaling 6,475.1 hours.  The release uses the same annotation models as the training case study: Gemma-4-E4B-it~\citep{google2026gemma4e4bit} for Camera and Qwen3.6-27B-FP8~\citep{qwen2026qwen36fp8} for Caption and Tag.  All three annotations are generated automatically under the versioned prompts and output schemas in \cref{app:annotation-schemas,app:annotation-prompts}.  Annotation-complete denotes that all three operations succeed and their outputs pass the corresponding schema checks.  Each released clip contains these annotations together with the quality measurements and duplicate relations produced during data processing.  \Cref{tab:vidaforge3m-overview} summarizes the released dataset.

This scales processing from the 200k-video pool used in the recipe study (\cref{sec:evaluation}) to 800k source videos from the same collection.  The release's source pool consists of the first 800,000 videos in source-shard order, without explicit category balancing.  The released semantic tags support content-based analysis and sampling.  Processing the full source collection is a natural direction for future expansion.

\begin{table}[h!]
    \centering
    \small
    \caption{\textsc{VidaForge-3M} at a glance. Statistics are computed over all 3,141,246 released clips.}
    \label{tab:vidaforge3m-overview}
    \begin{tabular}{@{}lr@{}}
        \toprule
        Property & Value \\
        \midrule
        Source videos processed & 800,000 \\
        Released scene-level clips & 3,141,246 \\
        Total duration & 6,475.1 hours \\
        Mean / median clip duration & 7.42 / 9.70 seconds \\
        Clip duration range & 1.0--10.0 seconds \\
        Video format & H.265/MP4 \\
        Video payload & 2.39 TB \\
        Release layout & 445 paired video/metadata shards \\
        \bottomrule
    \end{tabular}
\end{table}

\subsection{Large-Scale Data Processing Throughput}
\label{app:system-execution}

The VidaForge-3M run records the execution of every measured operation from
media conversion through annotation.  \Cref{tab:vidaforge3m-step-execution}
reports the workload, output, elapsed time, throughput, and hardware for each
step.

\begin{table}[h!]
    \centering
    \small
    \setlength{\tabcolsep}{2.8pt}
    \renewcommand{\arraystretch}{1.04}
    \caption{Execution of the VidaForge-3M pipeline.  Output entries report
    successful and failed items.  CPU steps use 800 Intel Xeon Gold 6530 cores;
    GPU steps use NVIDIA H200 GPUs.}
    \label{tab:vidaforge3m-step-execution}
    \begin{tabular}{@{}clrrrll@{}}
        \toprule
        Stage & Step & Input & Success / fail & Time & Throughput & Hardware \\
        \midrule
        I & Transcode & 800k videos & 800,000 / 0 & 5h00m & 44.36 videos/s & 800 CPU cores \\
        II & Detect & 800k videos & 800,000 / 0 & 5h27m & 40.74 videos/s & 800 CPU cores \\
        II & Clip & 800k videos & 3,144,490 / 0 & 12h42m & 68.71 clips/s & 800 CPU cores \\
        III & Context & 3.144M clips & 3,144,490 / 0 & 54m38s & 959.34 clips/s & 800 CPU cores \\
        III & Optical + Motion & 3.144M clips & 3,144,490 / 0 & 1h35m & 550.13 clips/s & 800 CPU cores \\
        III & Aesthetic & 3.144M clips & 3,144,490 / 0 & 2h26m & 357.15 clips/s & 32 H200 \\
        III & Text & 3.144M clips & 3,144,490 / 0 & 2h47m & 312.42 clips/s & 32 H200 \\
        III & PDQ Dedup & 3.144M clips & 3,144,490 / 0 & 13h40m & 63.90 clips/s & 800 CPU cores \\
        III & Cosmos Dedup & 3.144M clips & 3,144,490 / 0 & 2h47m & 313.22 clips/s & 32 H200 \\
        III & Select & 3.144M clips & 3,144,490 / 0 & 15m01s & 3,490.46 clips/s & 800 CPU cores \\
        IV & Camera & 3.144M clips & 3,144,408 / 82 & 18h34m & 47.01 clips/s & 24 H200 \\
        IV & Caption & 3.144M clips & 3,144,441 / 49 & 6h37m & 131.73 clips/s & 48 H200 \\
        IV & Tag & 3.144M clips & 3,141,399 / 3091 & 8h07m & 107.39 clips/s & 48 H200 \\
        \bottomrule
    \end{tabular}
\end{table}

\subsection{Media Composition}
\label{app:vidaforge3m-media}

\begin{figure}[h!]
    \centering
    \includegraphics[width=\textwidth]{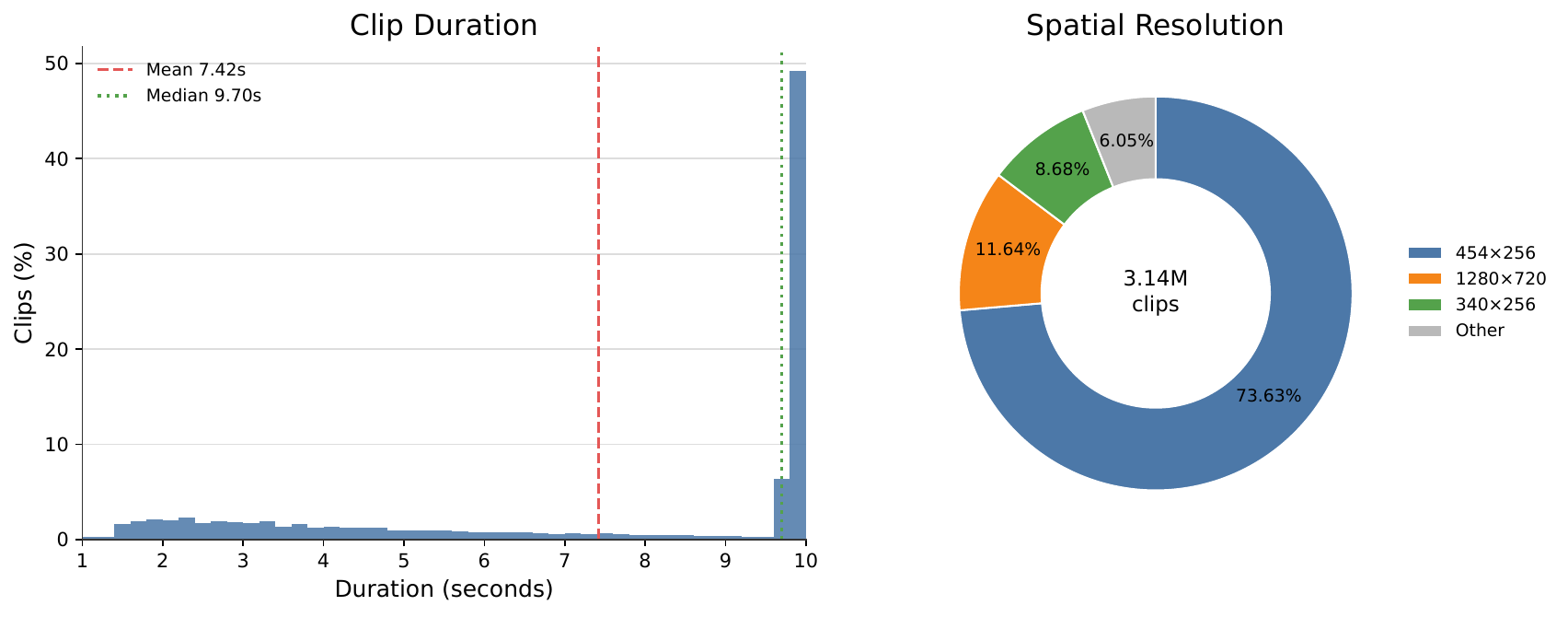}
    \caption{Media composition of \textsc{VidaForge-3M}. Left: clip-duration distribution over all released clips. Right: shares of the three most common spatial resolutions, with the remaining resolutions grouped as Other.}
    \label{fig:vidaforge3m-media-composition}
\end{figure}

\Cref{fig:vidaforge3m-media-composition} summarizes the temporal and spatial composition of \textsc{VidaForge-3M}.  Clip durations range from 1 to 10 seconds, with a mean of 7.42 seconds and a median of 9.70 seconds.  The most common resolution is $454\times256$ (73.63\%), followed by $1280\times720$ (11.64\%) and $340\times256$ (8.68\%); the remaining resolutions account for 6.05\%.  This distribution reflects VidaForge's standardization policy, which downsamples high-resolution inputs while leaving lower-resolution inputs at their original spatial scale.

\subsection{Curation Signals and Duplicate Relations}
\label{app:vidaforge3m-curation}

\begin{figure}[h!]
    \centering
    \includegraphics[width=\textwidth]{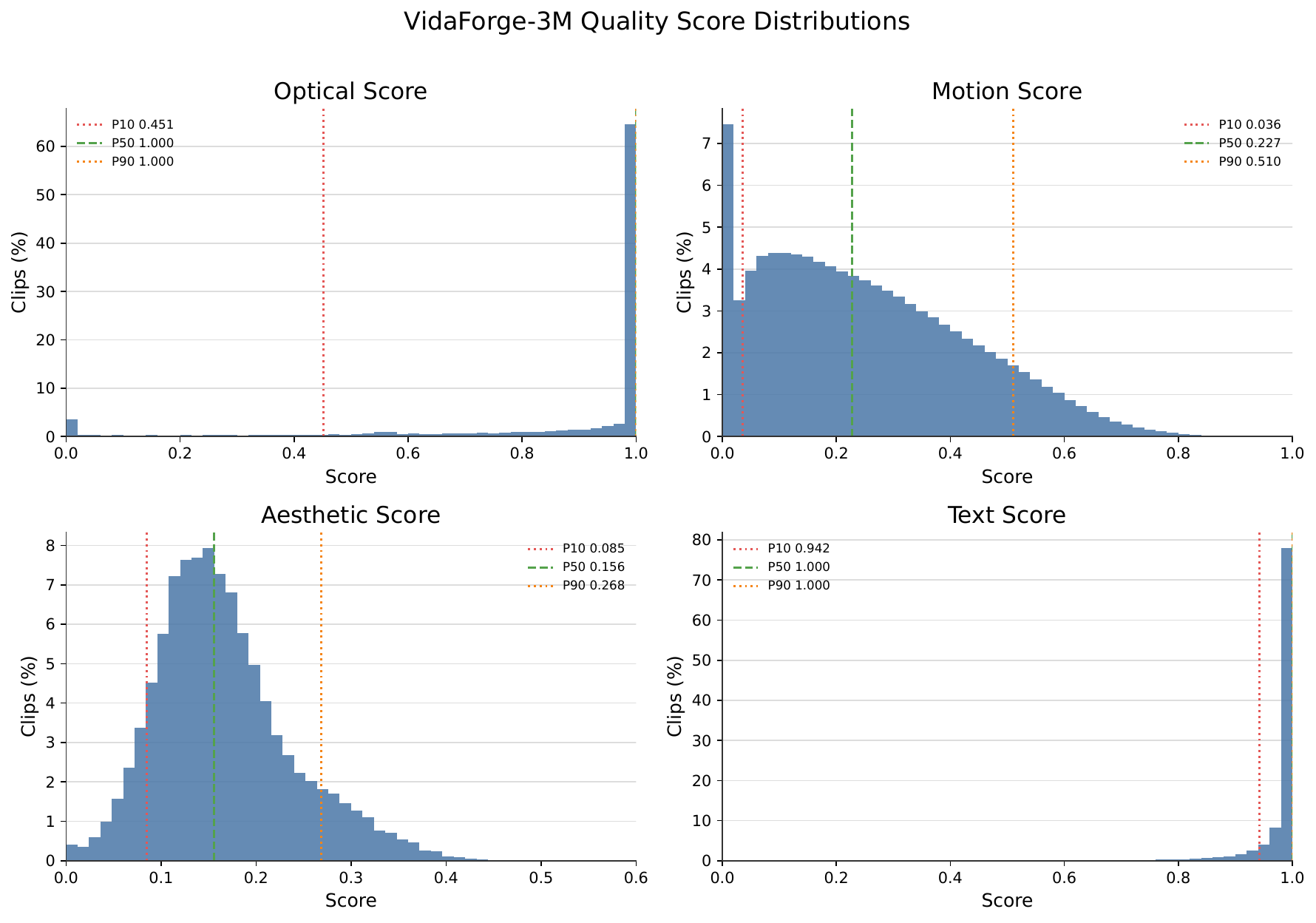}
    \caption{Quality-score distributions across all released \textsc{VidaForge-3M} clips. Vertical lines mark the 10th, 50th, and 90th percentiles for each measurement.}
    \label{fig:vidaforge3m-quality}
\end{figure}

\textsc{VidaForge-3M} retains every annotation-complete clip and publishes the signals needed to construct alternative curation recipes.  Each released clip records four normalized measurements: optical quality, motion, aesthetics, and visible text.  It also records separate duplicate-group assignments from PDQ and Cosmos-Embed.  The corresponding measurement and matching procedures are detailed in \cref{app:stage-selection}.  \Cref{fig:vidaforge3m-quality} shows the score distributions across the full release.  Optical scores concentrate near 1.  Text scores do as well, indicating that detected text occupies little image area in most clips.  Motion and aesthetic scores are distributed more broadly across their observed ranges.  Publishing the continuous values supports recipes based on thresholds, percentiles, or stratified sampling over the same processed clip population.

\Cref{tab:vidaforge3m-dedup} summarizes non-singleton duplicate groups within the released set.  PDQ~\citep{meta2019pdq} and Cosmos-Embed~\citep{nvidia2025cosmosembed1} provide perceptual and semantic relations, respectively.

\begin{table}[h!]
    \centering
    \small
    \caption{Non-singleton duplicate-group statistics within \textsc{VidaForge-3M}. Clip shares are computed over the full release; group-size percentiles are computed over non-singleton groups.}
    \label{tab:vidaforge3m-dedup}
    \begin{tabular}{@{}lrrr@{}}
        \toprule
        Method & Groups & Clips (share) & Size P50/P95 \\
        \midrule
        PDQ & 74,508 & 185,387 (5.90\%) & 2 / 3 \\
        Cosmos-Embed & 262,498 & 714,089 (22.73\%) & 2 / 3 \\
        \bottomrule
    \end{tabular}
\end{table}

The two group relations may overlap, so their clip shares are reported independently.  The public release retains clips from every duplicate group.  Each clip records its group identity, group size, and designated representative for each method.  A recipe can keep one representative, cap the number of clips per group, or retain the complete group without rerunning duplicate matching.  Together with the continuous measurements, these fields support multiple thresholding and deduplication recipes over the same release.

\clearpage
\subsection{Annotation Coverage and Granularity}
\label{app:vidaforge3m-annotations}

Every released clip includes Camera, Caption, and Tag annotations in the published schemas.  \Cref{tab:vidaforge3m-annotation-coverage} summarizes the released annotation interfaces; their complete schemas and prompts appear in \cref{app:annotation-schemas,app:annotation-prompts}.

\begin{table}[h!]
    \centering
    \small
    \caption{Annotation coverage in \textsc{VidaForge-3M}. Coverage is computed over all 3,141,246 released clips.}
    \label{tab:vidaforge3m-annotation-coverage}
    \begin{tabular}{@{}llr@{}}
        \toprule
        Annotation & Released structure & Coverage \\
        \midrule
        Camera & 21 structured motion fields & 3,141,246 (100\%) \\
        Caption & Four levels of increasing detail & 3,141,246 (100\%) \\
        Tag & Seven semantic taxonomy fields & 3,141,246 (100\%) \\
        \bottomrule
    \end{tabular}
\end{table}

\begin{figure}[h!]
    \centering
    \includegraphics[width=0.48\textwidth,trim={320pt 405pt 0pt 30pt},clip]{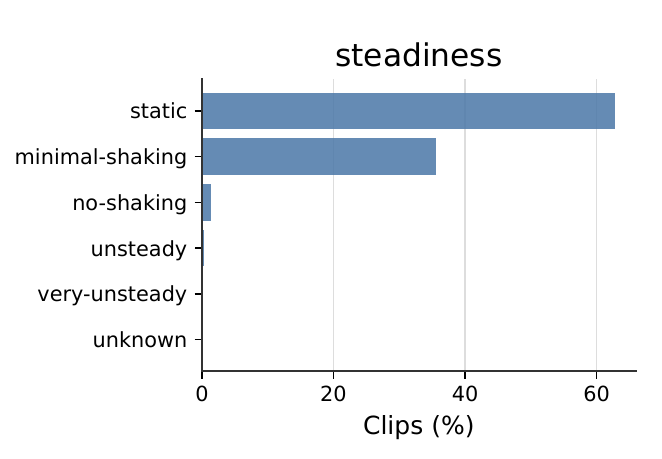}\hfill
    \includegraphics[width=0.48\textwidth,trim={320pt 210pt 0pt 245pt},clip]{figures/external/plot/outputs/vidaforge3m_statistics/camera_overview_distributions.pdf}
    \caption{Distributions of two complementary high-level annotations in \textsc{VidaForge-3M}. Steadiness describes camera stability; scene dynamics describes temporal activity in the visible content.}
    \label{fig:vidaforge3m-camera-high-level}
\end{figure}

The Camera schema separates camera stability from temporal activity within the scene.  In \cref{fig:vidaforge3m-camera-high-level}, 62.80\% of clips are labeled as having a static camera and 35.53\% as having minimal shaking.  Scene content is more varied: 47.84\% of clips are labeled dynamic, 43.91\% mostly static, and 8.25\% static.  Together, steadiness and scene dynamics allow sampling by camera stability and content motion separately.

\begin{figure}[p]
    \centering
    \includegraphics[width=\textwidth]{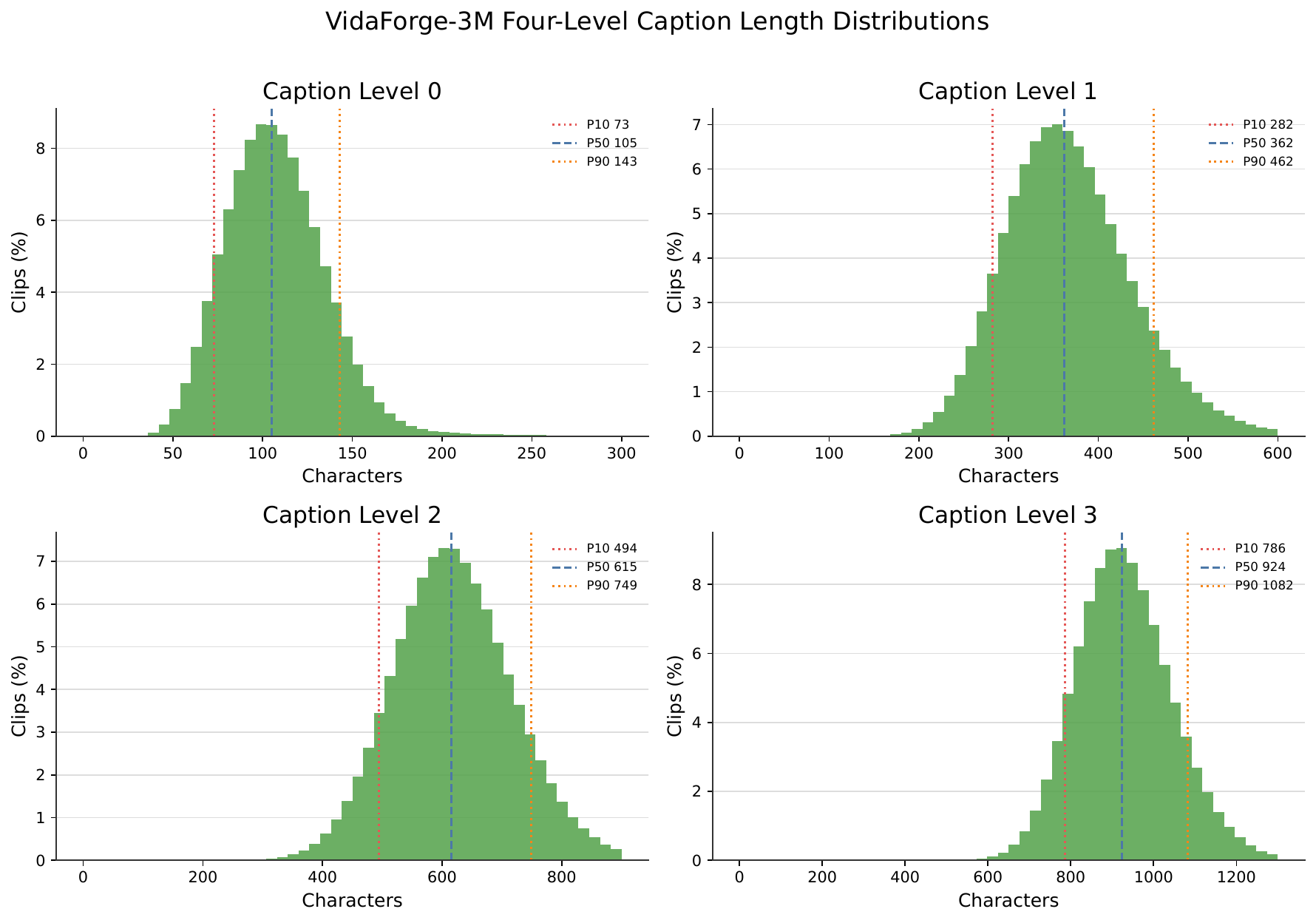}
    \caption{Character-length distributions of the four caption levels in \textsc{VidaForge-3M}. Vertical lines mark the 10th, 50th, and 90th percentiles.}
    \label{fig:vidaforge3m-caption-lengths}
\end{figure}

The four caption levels expose progressively richer descriptions of the same clip.  Their median lengths are 105, 362, 615, and 924 Unicode characters, respectively (\cref{fig:vidaforge3m-caption-lengths}).  A packaging recipe can therefore choose the amount of text supplied to a target training system without rerunning annotation.

\begin{figure}[p]
    \centering
    \includegraphics[width=\textwidth]{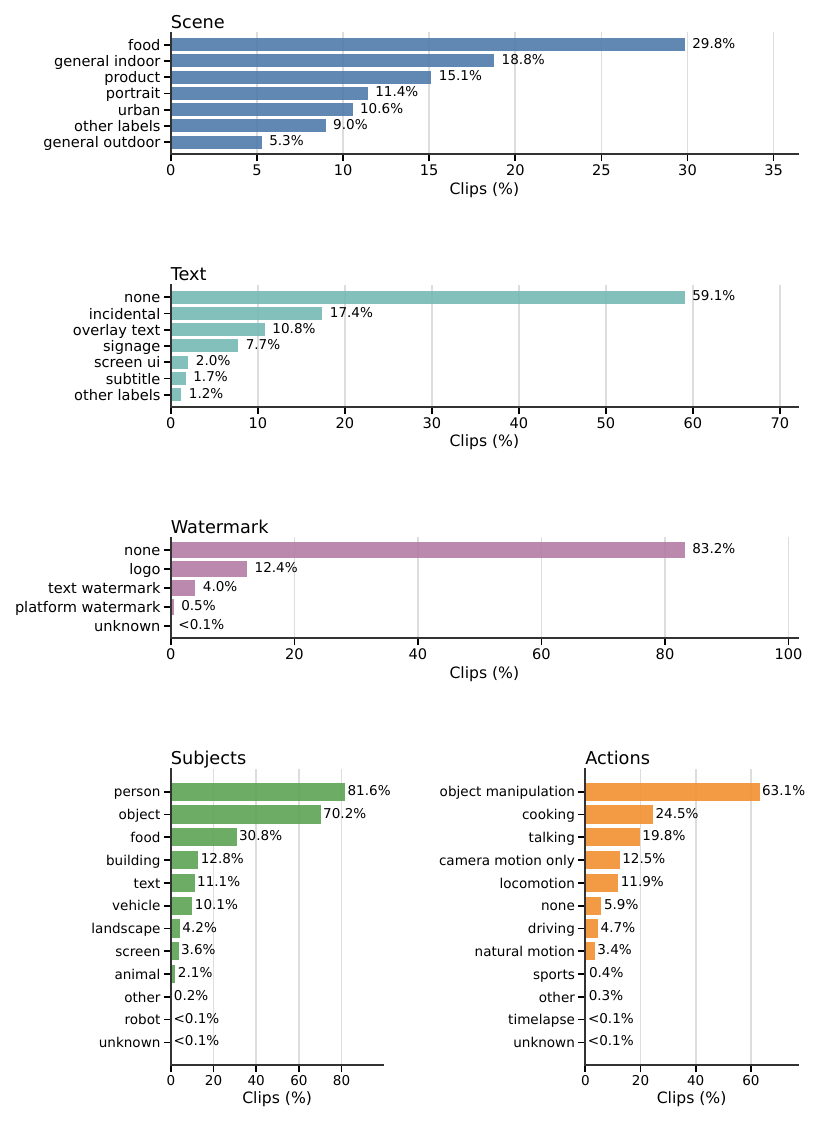}
    \caption{Selected Tag distributions in \textsc{VidaForge-3M}. Scene, Text, and Watermark are single-label fields; Subjects and Actions are multi-label fields, so percentages within either multi-label panel need not sum to 100\%.}
    \label{fig:vidaforge3m-tag-overview}
\end{figure}

The seven Tag fields provide a categorical view over domain, scene, subjects, actions, visual style, visible text, and watermark status.  Across the release, 94.2\% of clips receive the real-world domain label and 75.3\% receive the documentary style label.  Among the fields shown in \cref{fig:vidaforge3m-tag-overview}, 29.8\% receive the food scene label, 40.9\% receive a visible-text label other than none, and 16.8\% receive a watermark label other than none.  The multi-label person and object labels occur on 81.6\% and 70.2\% of clips, while object manipulation, cooking, and talking occur on 63.1\%, 24.5\%, and 19.8\%, respectively.  These distributions provide explicit axes for stratified sampling and quota-based data recipes.

\FloatBarrier
\section{Recipe Construction and Distribution Analysis}
\label{app:recipe_interventions}

\subsection{From the Processed Pool to Recipe Variants}
\label{app:recipe-flow}

The experiment begins from the same source snapshot and follows one
Ingestion and Segmentation path.  Processing 200k standardized source videos
produces 716k scene-level clips.  Stage~3 admits 352k clips and rejects
364k; \cref{fig:recipe-flow} records the primary rejection reason attached
to every rejected clip.  Optical quality accounts for 45.9\% of primary
rejections, motion for 20.7\%, semantic duplication for 19.2\%, aesthetics for
12.4\%, perceptual duplication for 1.8\%, and visible text for 0.07\%.
Because the complete decision record is retained, 70k rejected clips can
also be identified as matching more than one rejection condition.

Three 10k-clip validation sets are drawn from the resulting pools.  Removing
from training every clip that shares a parent \texttt{video\_id} with any
validation clip leaves 580k clips for training: 290k Selected and 290k
Rejected.  Mixed processes the full set once; Selected and Rejected process
their corresponding half twice. Random draws an independent random subset
of approximately 290k clips from Mixed for each run, retaining approximately
145k Selected and 145k Rejected clips. Each subset is used for two epochs,
keeping the number of processed clips fixed within each training system.

\subsection{Selection Scores, Redundancy, and Coverage}
\label{app:recipe-population-shift}

In this study, coverage refers to the number of different clips encountered
during training and the regions those clips occupy in the map of semantic
video embeddings generated by Cosmos-Embed.  Quality is represented by the four selection measurements and
within-dataset redundancy derived from the duplicate groups. Under the same budget of
approximately 580k processed clips, Mixed uses approximately 580k different
clips, while Selected, Random, and Rejected each use approximately 290k
different clips for two passes through their respective datasets.

\begin{figure}[p]
  \centering
  \includegraphics[width=\textwidth]{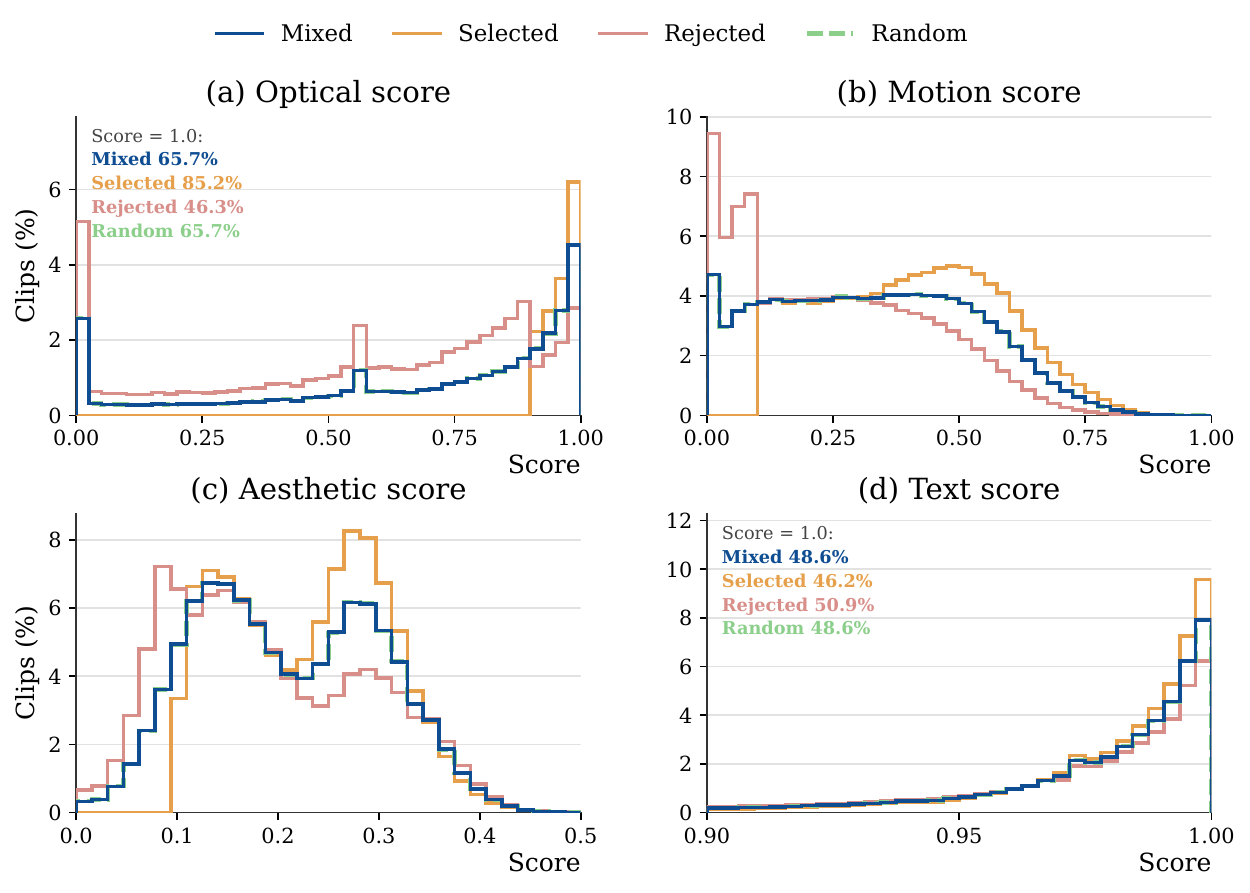}
  \caption{Distributions of the four selection scores in the final Mixed,
  Selected, and Rejected training datasets.  Each histogram is normalized by
  the number of valid clips in its dataset.  Optical and text scores equal to
  1.0 are omitted from the histogram bins and reported separately inside their
  panels; they remain in the normalization denominator.
  The visible-text panel focuses on scores $\geq 0.9$ for readability.}
  \label{fig:recipe-score-distributions}
\end{figure}

\begin{figure}[p]
  \centering
  \includegraphics[width=\textwidth]{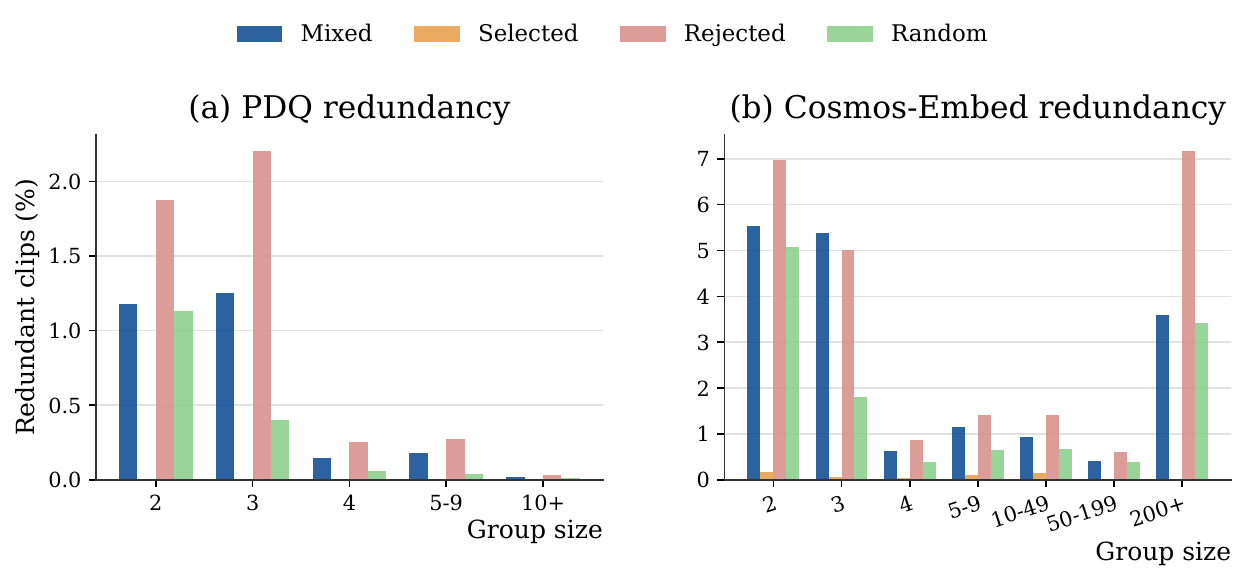}
  \caption{PDQ and semantic redundancy by within-dataset group size. A
  group of size $n$ contributes $n-1$ redundant clips; each bar is reported as
  a percentage of the corresponding final training dataset.}
  \label{fig:recipe-duplicate-distributions}
\end{figure}

The recorded selection measurements confirm that the two component pools
differ along the dimensions used by the recipe.  Mean motion changes from
0.252 for Rejected to 0.408 for Selected, mean optical score from 0.777 to
0.994, and mean aesthetic score from 0.191 to 0.230.  Within the final training
datasets, PDQ redundancy is 2.77\%/0.00\%/4.63\% and semantic redundancy
is 17.64\%/0.53\%/23.44\% for Mixed/Selected/Rejected. Redundancy counts
$n-1$ clips per group of size $n$, divided by the number of valid clips in
the dataset. Groups are restricted to the clips present in each dataset.
Random has mean PDQ and semantic redundancy of 1.64\% and 12.41\%,
respectively. Subsampling reduces the number of clips present together in
duplicate groups; groups with only one remaining clip contribute no redundancy.
This explains Random's lower redundancy than Mixed without additional
deduplication. Its quality scores and redundancy in \cref{fig:recipe-quality}
are computed separately for each of the 3 sampled subsets, then averaged.
The complete score and duplicate-group distributions for Mixed, Selected,
and Rejected appear in
\cref{fig:recipe-score-distributions,fig:recipe-duplicate-distributions}.

Selection changes semantic composition across the full candidate pool.  We
compute semantic video embeddings with Cosmos-Embed for all clips and visualize a deterministic
subset with t-SNE~\citep{vandermaaten2008visualizing}.  We reduce the embeddings to 50 principal components and run
t-SNE with perplexity 50 for 1,000 iterations.  \Cref{fig:recipe-semantic-coverage} colors
the pooled 95\% support by smoothed local membership dominance.  Coherent
Selected- and Rejected-dominant regions show how Stage~3 changes semantic
composition.  Mixed retains both sets of regions and spans their combined
semantic coverage.

\FloatBarrier
\section{Training and Evaluation Details}
\label{app:training}

\subsection{Controls and Validation Isolation}
\label{app:training-controls}

\begin{table}[t]
  \centering
  \small
  \caption{Training and evaluation counts for the reported study.
  Each loss evaluation uses one checkpoint and one validation set.}
  \label{tab:experiment-scale}
  \begin{tabular}{@{}lll@{}}
    \toprule
    Model & Experiment & Count \\
    \midrule
    Wan + V-JEPA~2 & From-scratch pretraining & 24 runs \\
    Wan & Validation loss & 252 evaluations \\
    V-JEPA~2 & Validation loss & 720 evaluations \\
    Wan & VBench & 36 evaluations \\
    V-JEPA~2 & SSv2 & 60 evaluations \\
    V-JEPA~2 & Kinetics-400 & 4 evaluations \\
    \bottomrule
  \end{tabular}
  \par\smallskip
  \begin{minipage}{0.94\linewidth}
  \footnotesize Each model uses 4 recipes and 3 pretraining runs.
  Loss counts cover 3 validation sets at 7 checkpoints per Wan run and
  20 per V-JEPA~2 run. VBench evaluates 3 checkpoints per run;
  SSv2 evaluates 5 checkpoints per run; Kinetics-400
  evaluates one final checkpoint per recipe. Each downstream probe
  evaluation trains 3 learning-rate variants for 20 epochs, totaling
  180 SSv2 and 12 Kinetics-400 probes.
  \end{minipage}
\end{table}

Wan~2.1 and V-JEPA~2 use the same Mixed, Selected, Random, and Rejected recipe
definitions. All runs start from scratch. Both models use 3 runs per recipe,
with training seeds matched across Mixed, Selected, and Rejected.
The architectures, optimizer configurations, data-loading policies,
scheduled updates, and evaluation inputs remain fixed within a model family.
The approximately 580k Mixed dataset is traversed once, while the approximately
290k Selected, Random, and Rejected datasets are traversed twice.
For Random, each run uses an independently sampled subset with approximately
equal numbers of Selected and Rejected clips. The subset stays fixed during
that run. Random's standard deviation therefore includes both subset-sampling
and pretraining variability; the other recipes retain fixed datasets across seeds.

Three 10k-clip validation sets are sampled from the Selected, Mixed, and
Rejected pools.  Their parent
\texttt{video\_id} values are excluded from all four training recipes, so no
training clip shares a source video with any validation clip.  After
distributed drop-last, both model families evaluate each validation set at
every checkpoint.

\subsection{Wan~2.1 Protocol and Diagnostics}
\label{app:wan-details}

VidaForge packages each recipe as the tensor caches consumed by the
NeMo-AutoModel~\citep{nvidia2026nemoautomodel} Wan training path.  We train Wan~2.1-1.3B from scratch with
global batch size 96 for 5,950 optimizer updates. We use AdamW with
$\beta=(0.9,0.95)$, $\epsilon=10^{-8}$, and constant weight decay 0.1.
The learning rate follows cosine decay from $5\times10^{-5}$ to $10^{-6}$
over 5,950 updates, with no warmup. The optimizer configuration, bucketed
loading policy, and checkpoint schedule are shared across runs.
Temporal buckets contain 49, 73, 97, 125, 149, 197, or 249 frames, with
spatial dimensions varying by bucket. During packaging, frames are uniformly
sampled over the full clip duration according to its temporal bucket and
encoded into latent caches; training reads these precomputed caches.

\begin{figure}[p]
  \centering
  \includegraphics[width=0.9\textwidth]{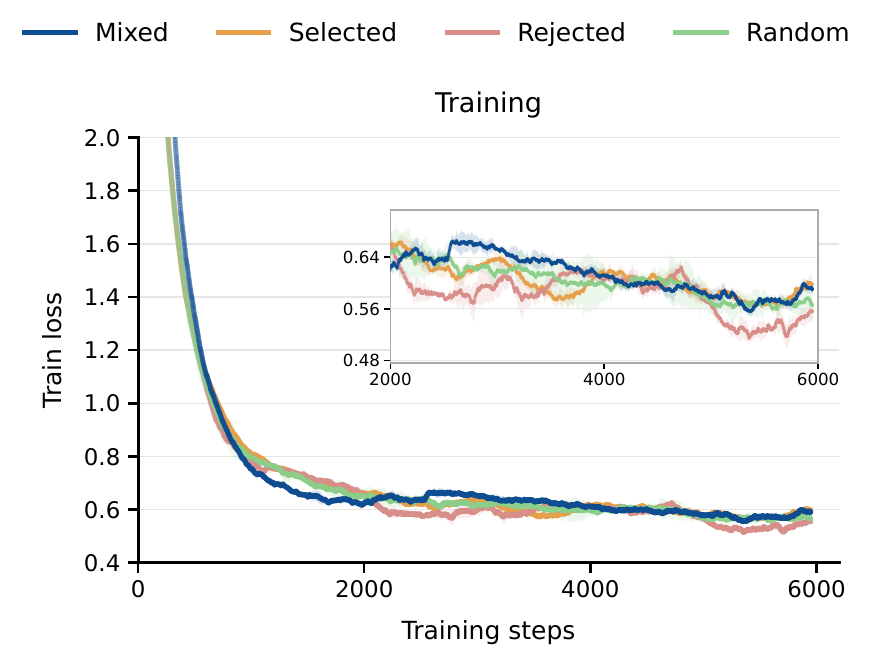}
  \caption{Wan~2.1 flow-matching training loss for all four recipes across 3 runs
  (mean $\pm$ std). The inset enlarges the later training phase.}
  \label{fig:wan-training-loss}
\end{figure}

Training loss is the per-update flow-matching loss logged every two updates.
\Cref{fig:wan-training-loss,fig:wan-losses} summarize training and validation
loss for all four recipes across 3 runs
(mean $\pm$ std). Validation loss is unsmoothed.
Across all evaluated checkpoints and validation sets, average cross-seed std
is 0.0038 for Mixed, 0.0031 for Selected, and 0.0065 for Rejected.
These small values explain the narrow shaded bands in the validation curves.

VBench~\citep{huang2024vbench} evaluation uses the final checkpoint at update 5,950 and
generates 3 videos per official prompt.\footnote{One Random evaluation replaces
an invalid video with a copy of another generated video for the same prompt.}
All runs use the same
initial noise seeds, sampler settings, resolution, frame count, and frame rate.
The official Quality, Semantic, and
Total scores summarize the 16 dimensions. As a sensitivity analysis,
we also recompute Quality and Total without Dynamic Degree for all four recipes.
Mixed has the highest mean scores
with or without Dynamic Degree, showing that its VBench advantage extends
beyond this motion metric. \Cref{tab:wan-vbench-final-three-seed} reports
the aggregates, and \cref{tab:wan-vbench-dimensions} reports each dimension.

\begin{figure}[p]
  \centering
  \includegraphics[width=0.95\linewidth]{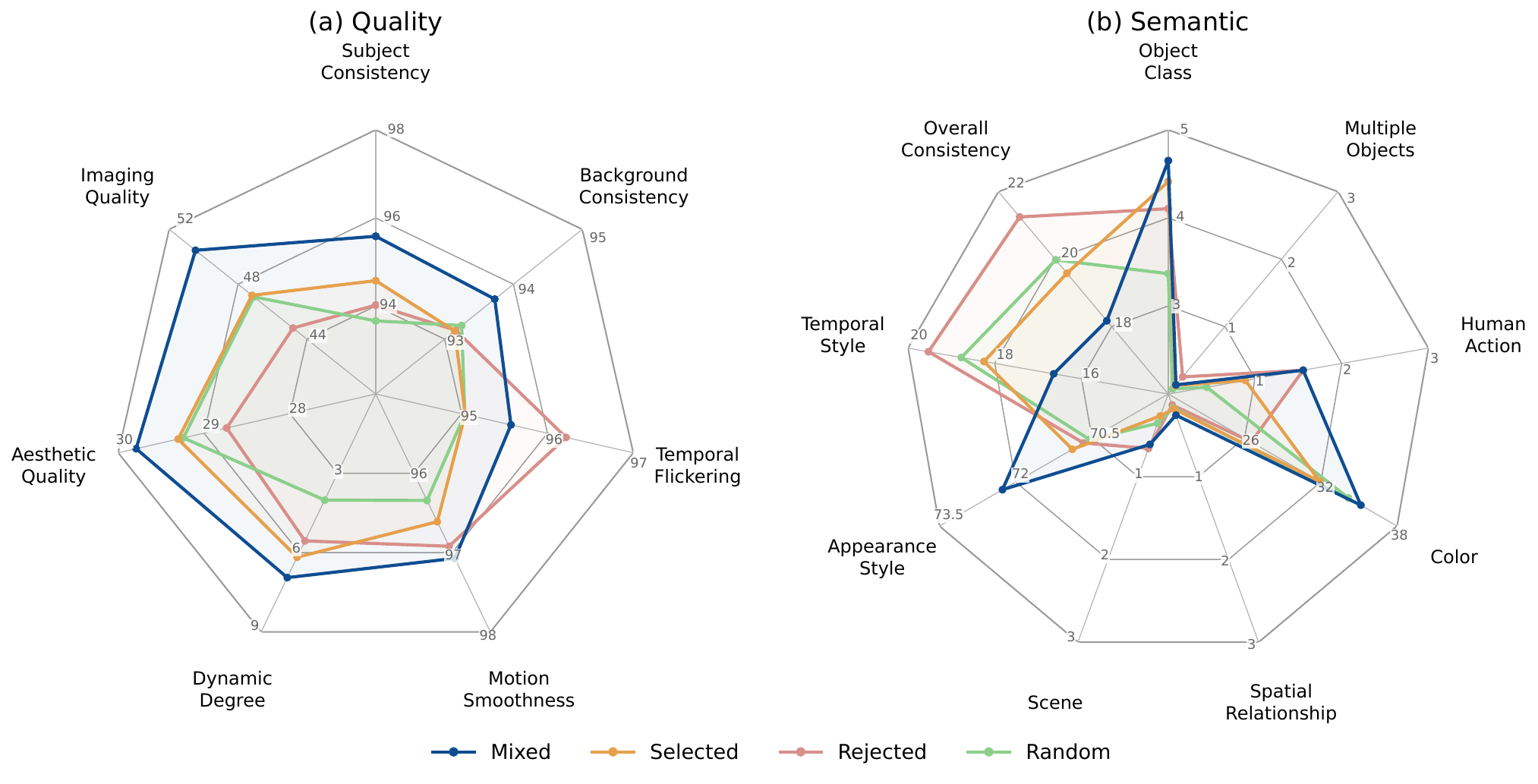}
  \caption{VBench dimension scores for Wan~2.1-1.3B, averaged over 3 runs.
  Left: Quality dimensions. Right: Semantic dimensions.
  Axes are rescaled to highlight differences;
  \cref{tab:wan-vbench-dimensions} gives per-dimension mean $\pm$ std.}
  \label{fig:wan-vbench}
  \label{fig:wan-vbench-radar}
\end{figure}

\Cref{fig:wan-vbench} visualizes the dimension means for all four recipes.
Mixed has the highest mean scores in Imaging Quality, Aesthetic Quality,
Color, Subject Consistency, and Appearance Style. Selected improves Imaging
Quality over Rejected, but remains behind Mixed.

\begin{table}[htbp]
  \centering
  \caption{Final VBench aggregate scores for all four recipes across 3 Wan runs
  (mean $\pm$ std). Total$^{\dagger}$ excludes Dynamic Degree. Higher is better.}
  \label{tab:wan-vbench-final-three-seed}
  \small
  \setlength{\tabcolsep}{3.5pt}
  \begin{tabular}{@{}lcccc@{}}
    \toprule
    Recipe & Quality & Semantic & Total & Total$^{\dagger}$ \\
    \midrule
    Mixed & $\mathbf{71.65}\pm0.74$ & $\mathbf{16.60}\pm1.25$ & $\mathbf{60.64}\pm0.83$ & $\mathbf{64.95}\pm1.52$ \\
    Selected & $70.61\pm0.51$ & $16.28\pm0.81$ & $59.74\pm0.55$ & $64.04\pm0.74$ \\
    Random & $70.25\pm0.80$ & $16.43\pm0.76$ & $59.48\pm0.67$ & $63.90\pm0.64$ \\
    Rejected & $70.25\pm1.19$ & $16.08\pm0.89$ & $59.42\pm1.06$ & $63.73\pm1.47$ \\
    \bottomrule
  \end{tabular}
\end{table}

\begin{table}[htbp]
  \centering
  \caption{Final VBench Total for Mixed, Selected, and Rejected under each Wan
  training seed. Mixed ranks first among these three under all 3 seeds;
  the ordering of Selected and Rejected varies.}
  \label{tab:wan-vbench-seeds}
  \small
  \begin{tabular}{@{}lrrr@{}}
    \toprule
    Training seed & Mixed & Selected & Rejected \\
    \midrule
    Seed 1 & \textbf{60.95} & 59.81 & 59.50 \\
    Seed 2 & \textbf{59.69} & 59.16 & 58.32 \\
    Seed 3 & \textbf{61.27} & 60.26 & 60.43 \\
    \bottomrule
  \end{tabular}
\end{table}

\begin{table}[htbp]
  \centering
  \caption{Normalized VBench dimension scores for all four recipes at the final Wan checkpoint
  across 3 runs (mean $\pm$ std). Higher is better; bold marks
  the highest displayed mean in each row.}
  \label{tab:wan-vbench-dimensions}
  \small
  \setlength{\tabcolsep}{3.5pt}
  \begin{tabular}{@{}lcccc@{}}
    \toprule
    Dimension & Mixed & Selected & Random & Rejected \\
    \midrule
    \multicolumn{5}{l}{\textit{Quality dimensions}} \\
    Subject Consistency & $\mathbf{95.58}\pm2.40$ & $94.57\pm1.73$ & $93.66\pm2.08$ & $94.02\pm2.73$ \\
    Background Consistency & $\mathbf{93.73}\pm0.75$ & $93.15\pm1.58$ & $93.25\pm1.32$ & $93.16\pm1.10$ \\
    Temporal Flickering & $95.58\pm2.38$ & $95.04\pm0.66$ & $95.04\pm0.96$ & $\mathbf{96.22}\pm1.33$ \\
    Motion Smoothness & $\mathbf{97.07}\pm1.96$ & $96.61\pm0.55$ & $96.34\pm1.32$ & $96.92\pm1.36$ \\
    Dynamic Degree$^{\dagger}$ & $\mathbf{6.94}\pm9.63$ & $6.17\pm2.38$ & $4.01\pm1.75$ & $5.56\pm5.26$ \\
    Aesthetic Quality & $\mathbf{29.79}\pm0.87$ & $29.30\pm0.25$ & $29.24\pm0.75$ & $28.74\pm1.13$ \\
    Imaging Quality & $\mathbf{50.47}\pm2.17$ & $47.19\pm0.55$ & $47.08\pm2.82$ & $44.81\pm2.83$ \\
    \midrule
    \multicolumn{5}{l}{\textit{Semantic dimensions}} \\
    Object Class & $\mathbf{4.65}\pm0.37$ & $4.41\pm0.24$ & $3.37\pm0.95$ & $4.11\pm0.63$ \\
    Multiple Objects & $0.14\pm0.13$ & $0.13\pm0.07$ & $0.08\pm0.03$ & $\mathbf{0.25}\pm0.18$ \\
    Human Action & $\mathbf{1.56}\pm0.69$ & $0.89\pm0.38$ & $0.44\pm0.19$ & $\mathbf{1.56}\pm0.84$ \\
    Color & $\mathbf{35.15}\pm12.18$ & $31.93\pm5.89$ & $34.17\pm7.53$ & $26.53\pm8.79$ \\
    Spatial Relationship & $\mathbf{0.26}\pm0.20$ & $0.18\pm0.09$ & $0.17\pm0.07$ & $0.13\pm0.06$ \\
    Scene & $0.61\pm0.43$ & $0.27\pm0.20$ & $0.35\pm0.05$ & $\mathbf{0.66}\pm0.26$ \\
    Appearance Style & $\mathbf{72.26}\pm0.65$ & $70.89\pm0.30$ & $70.55\pm1.76$ & $70.67\pm1.21$ \\
    Temporal Style & $16.64\pm0.65$ & $18.25\pm1.96$ & $18.77\pm0.43$ & $\mathbf{19.53}\pm1.05$ \\
    Overall Consistency & $18.17\pm1.54$ & $19.58\pm1.09$ & $19.98\pm0.51$ & $\mathbf{21.25}\pm1.57$ \\
    \bottomrule
  \end{tabular}
  \par\smallskip
  \begin{minipage}{0.96\linewidth}
    \footnotesize $^{\dagger}$Dynamic Degree measures motion presence and
    is included in the official aggregate.
  \end{minipage}
\end{table}

\subsection{V-JEPA~2-1B Protocol and Diagnostics}
\label{app:vjepa-details}

VidaForge packages each recipe as the manifest consumed by the official
V-JEPA~2 repository.  We train the 1B-parameter V-JEPA~2 encoder
from scratch with global batch size 256 for 2,280 updates using 3 runs per
recipe. Curves show mean $\pm$ std, where std is the sample standard deviation.
Training loss, validation loss, and SSv2 trajectories cover all four recipes.
Random varies both the sampled subset and training seed; final accuracies
appear in \cref{tab:recipe-final-results}.
Evaluation protocols are fixed across runs.

\begin{figure}[p]
  \centering
  \includegraphics[width=\textwidth]{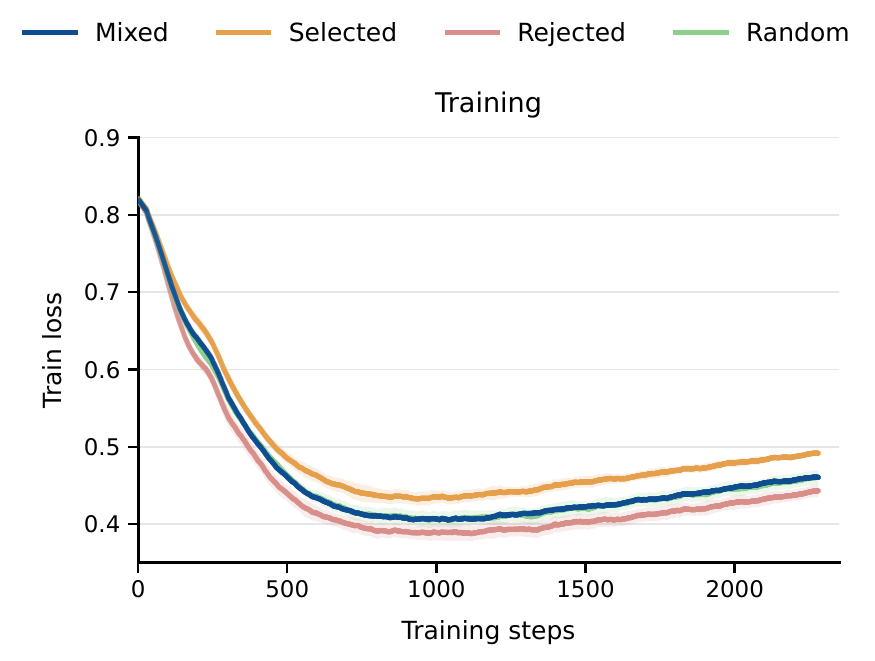}
  \caption{V-JEPA~2 masked-prediction training loss for all four recipes
  across 3 runs. Each run's curve is first smoothed;
  solid lines and shaded bands then show mean $\pm$ std.
  Rejected has the lowest final mean loss.}
  \label{fig:vjepa-training-loss}
\end{figure}

We evaluate encoder checkpoints on the three 10k-clip validation sets using
a fixed evaluation seed.
The training curve in
\cref{fig:vjepa-training-loss} and the validation curves in
\cref{fig:vjepa-comparison} reach their minima near the middle of the schedule
and subsequently rise, with Rejected retaining the lowest loss.  V-JEPA~2
predicts latent features generated by a target encoder that changes throughout
training as an exponential moving average of the context
encoder~\citep{assran2025vjepa2}.  Training uses the current target encoder,
while validation at each checkpoint uses the target encoder saved with that
checkpoint.  Losses across time therefore compare predictions against
different target features.  Because the target features evolve during training,
prediction loss can increase even as the learned representations improve.  Similar behavior
has been reported for V-JEPA and V-JEPA~2
training\footnote{\href{https://github.com/facebookresearch/jepa/issues/68}{V-JEPA issue 68};
\href{https://github.com/facebookresearch/vjepa2/issues/56}{V-JEPA~2 issue 56}.}.
Across the same interval, SSv2 accuracy continues to improve, separating the
non-monotonic pretraining objective from action-recognition utility in these
runs.

\paragraph{Validation-loss variability.}
Across checkpoints and validation sets, the median cross-seed standard
deviation is 0.0016 for Mixed, compared with 0.0064 for Selected and 0.0089
for Rejected. Selected and Rejected therefore show approximately $4\times$
and $5.5\times$ the validation-loss variability of Mixed.

Representation utility is measured by adapting the frozen attentive-probe
evaluator from the official V-JEPA~2 repository to the complete
Something-Something V2 benchmark~\citep{goyal2017something}.  We freeze the target
encoder, sample 16 frames with temporal step 3 at resolution 256, and train only
the probe for 20 epochs using approximately 169k training and 25k validation videos,
with 8 GPUs and local batch size 4. Data and probe seeds are fixed across runs.
Three independent probes are trained in parallel with initial learning rates
$10^{-3}$, $3\times10^{-4}$, and $10^{-4}$. Each uses AdamW with
$\beta=(0.9,0.999)$, $\epsilon=10^{-8}$, and constant weight decay 0.1.
Each learning rate follows cosine decay to zero over 20 epochs, with no warmup.
For each encoder checkpoint, the reported metric is validation
top-1 at probe epoch 20, taking the maximum across these probes. We evaluate
checkpoints throughout training and report final results at the end of pretraining.

\paragraph{Kinetics-400 evaluation.}
We additionally evaluate one final encoder per recipe on
Kinetics-400~\citep{kay2017kinetics} using frozen attentive probes.
The dataset manifests contain approximately 241k training and 20k validation videos.
Inputs contain 16 frames at source-frame step 4 and resolution 256, with
one temporal segment and one spatial view. We train probes for 20 epochs
with local batch size 16 on 8 GPUs (global batch size 128) and probe seed 0.
The three probes use the same AdamW settings, learning rates, cosine schedule,
and zero warmup as SSv2; learning rates are not scaled with the larger batch.
We report the highest validation top-1 across these probes
at epoch 20, following the same reporting rule as SSv2.
The loader retries failed reads by substituting another video and its label;
distributed sampling pads inputs to equal rank lengths.
\Cref{tab:recipe-final-results} reports the four completed runs.

\paragraph{SSv2 results across runs.}
\Cref{tab:ssv2-training-seeds} reports final accuracy for all four recipes
and gains over Selected and Rejected paired by pretraining seed.
Mixed ranks above Selected and Rejected under every matched seed.
Its mean gains are 0.543 and 1.157 percentage points, respectively.
All four recipes improve markedly in the latter part of pretraining.
Under the fixed probe protocol, standard deviations capture pretraining
variation for Mixed, Selected, and Rejected, and both subset-sampling and
pretraining variation for Random.

\begin{table}[ht]
  \centering
  \small
  \setlength{\tabcolsep}{5pt}
  \renewcommand{\arraystretch}{1.08}
  \caption{Final SSv2 top-1 accuracy (\%) and paired gains (percentage points)
  across 3 runs per recipe (mean $\pm$ std).
  Mixed, Selected, and Rejected use matched training seeds; Random varies
  both subset sampling and training seed. Paired gains use matched seeds
  before aggregation; displayed values are rounded.}
  \label{tab:ssv2-training-seeds}
  \begin{tabular}{@{}lrrrr@{}}
    \toprule
    Recipe / comparison & Run 1 & Run 2 & Run 3 & Mean $\pm$ std \\
    \midrule
    Mixed & 16.105 & 16.412 & 16.250 & $16.256\pm0.153$ \\
    Selected & 15.387 & 16.133 & 15.617 & $15.712\pm0.382$ \\
    Random & 15.535 & 15.213 & 15.185 & $15.311\pm0.194$ \\
    Rejected & 15.153 & 15.407 & 14.737 & $15.099\pm0.338$ \\
    \midrule
    Mixed--Selected & $+0.719$ & $+0.278$ & $+0.633$ & $+0.543\pm0.233$ \\
    Mixed--Rejected & $+0.953$ & $+1.005$ & $+1.513$ & $+1.157\pm0.309$ \\
    \bottomrule
  \end{tabular}
\end{table}

\subsection{Downstream Performance During Pretraining}
\label{app:downstream-dynamics}

\begin{figure}[htbp]
  \centering
  \includegraphics[width=0.68\textwidth]{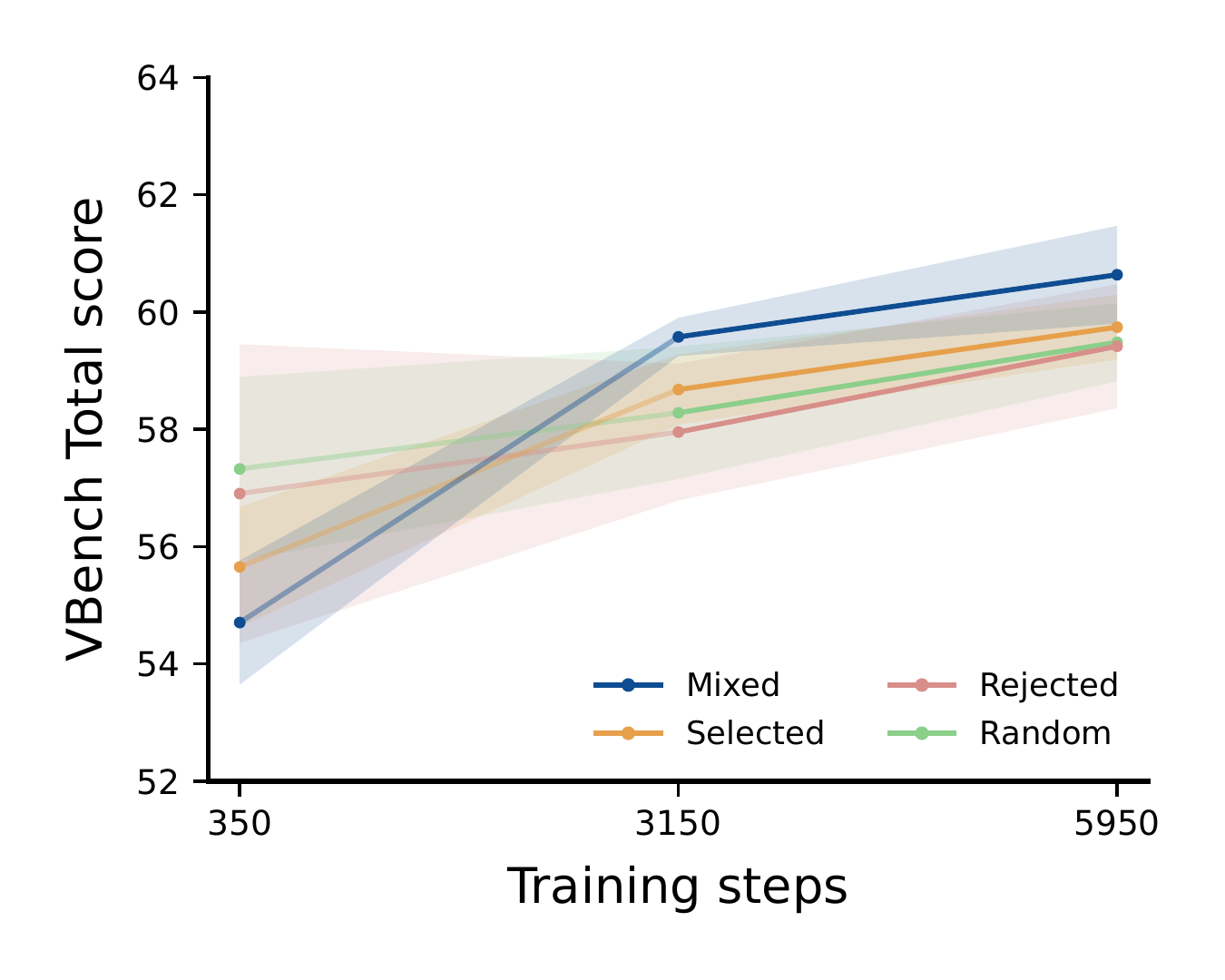}
  \caption{VBench Total for all four recipes during Wan pretraining
  (mean $\pm$ std across 3 runs). Random varies both subset sampling and
  training seed; the other recipes vary the training seed.}
  \label{fig:wan-vbench-trajectories}
\end{figure}

\begin{figure}[htbp]
  \centering
  \includegraphics[width=0.68\textwidth]{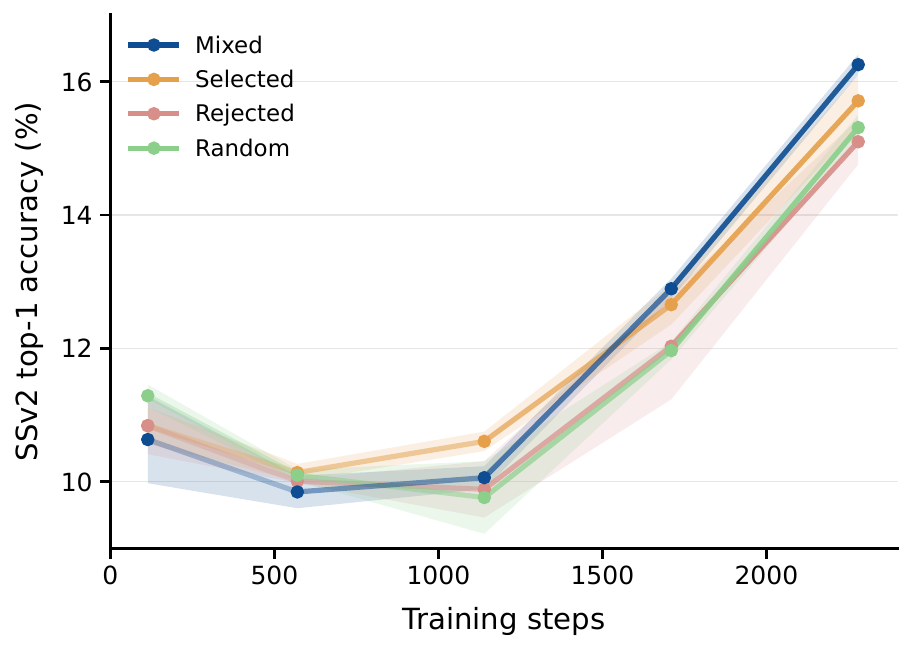}
  \caption{SSv2 frozen-probe accuracy for all four recipes
  during V-JEPA~2 pretraining (mean $\pm$ std across 3 runs).
  Random varies both subset sampling and training seed;
  the other recipes vary the training seed.}
  \label{fig:ssv2-trajectories}
\end{figure}

\Cref{fig:wan-vbench-trajectories,fig:ssv2-trajectories} show downstream
performance at intermediate and final checkpoints for all four recipes
across 3 runs. VBench evaluates Wan at updates 350, 3,150, and 5,950;
SSv2 evaluates V-JEPA~2 at updates 114, 570, 1,140, 1,710, and 2,280.
Mean VBench Total increases across the three evaluated checkpoints for
all four recipes, while SSv2 accuracy rises markedly in the latter part
of pretraining. The smaller subsets continue to improve during repeated
training, so their lower final scores than Mixed do not imply a decline
in downstream performance during the second pass. These trajectories
complement the final comparison in \cref{tab:recipe-final-results}.

\FloatBarrier
\section{Qualitative Examples}
\label{app:qualitative-examples}

This appendix illustrates the selection rules and annotations used in the
coverage--quality study. Each example pairs video frames with recorded scores,
decisions, or annotations.

\subsection{Selection Examples}
\label{app:selection-examples}

\begin{figure}[htbp]
  \centering
  \includegraphics[width=0.95\linewidth]{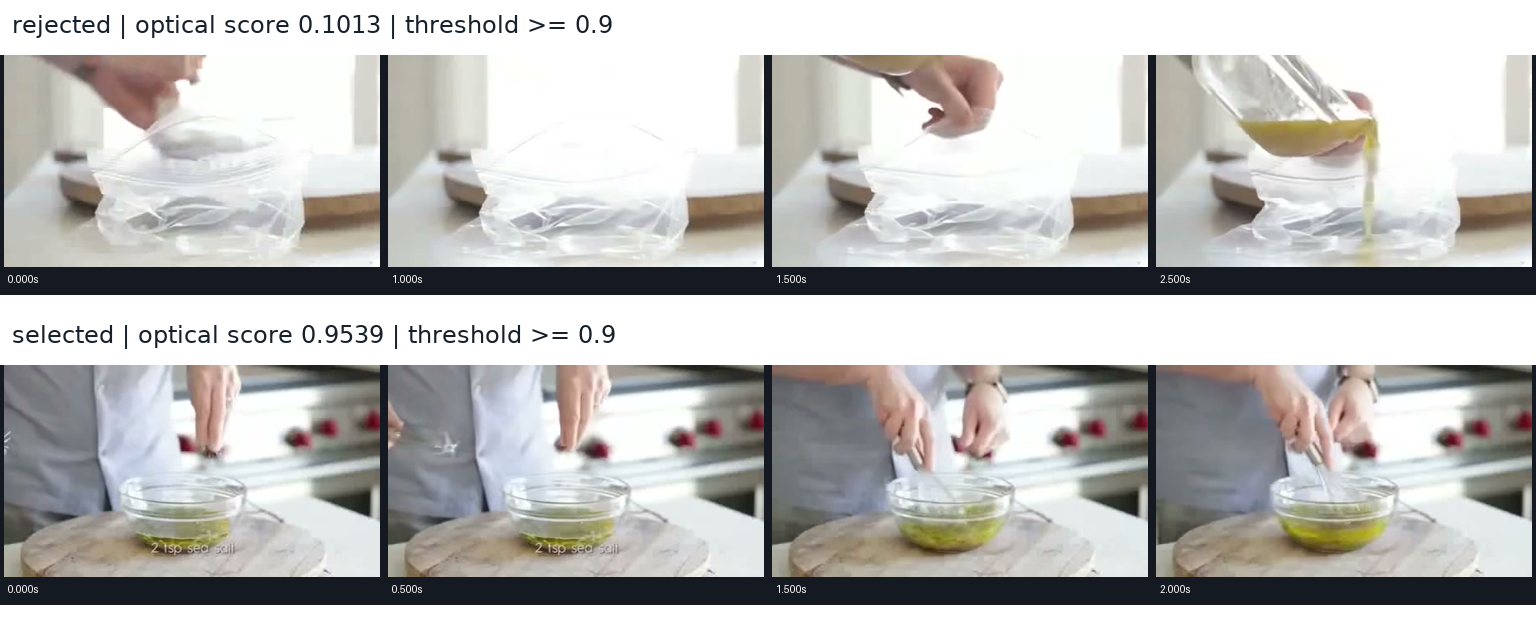}
  \caption{\textbf{Optical filtering.} Two clips from the same cooking video
  illustrate optical filtering. Overexposed regions cause the first clip to
  fail the 0.9 threshold; the second clip passes.}
  \label{fig:example-optical}
\end{figure}

\begin{figure}[htbp]
  \centering
  \includegraphics[width=0.95\linewidth]{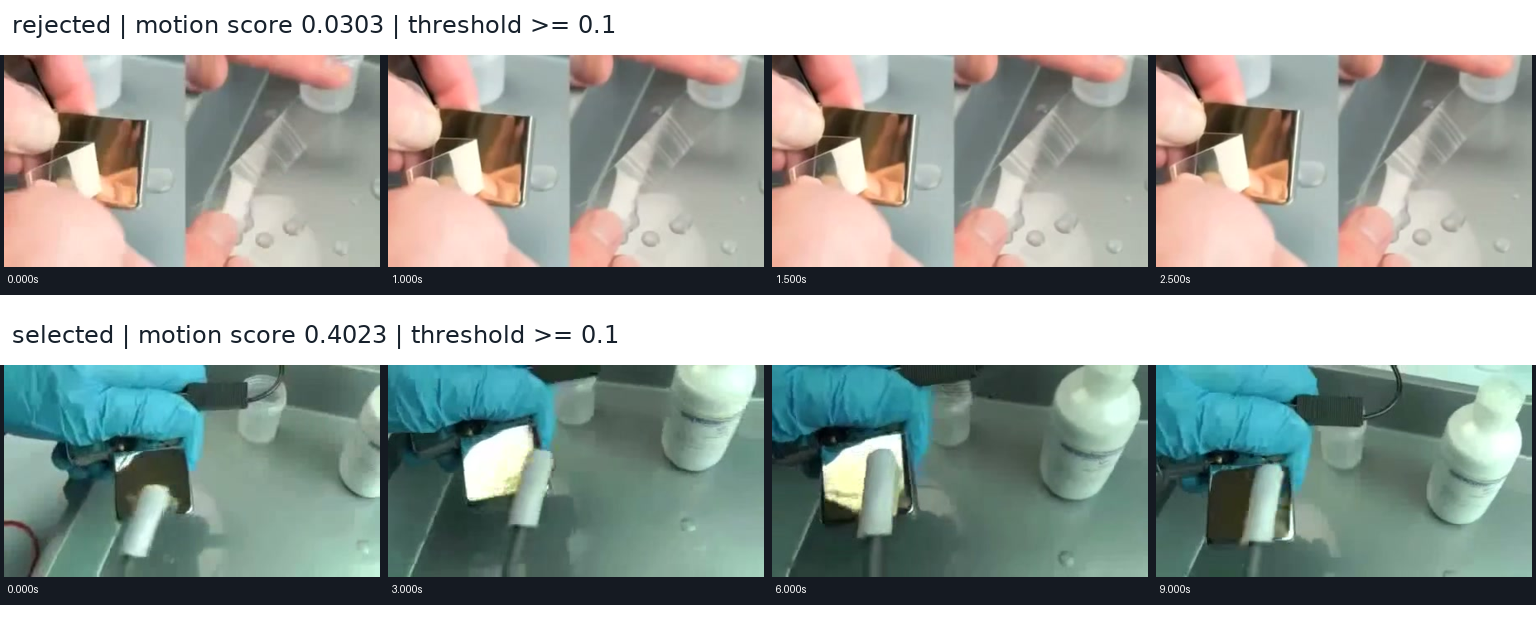}
  \caption{\textbf{Motion filtering.} Two clips from the same source video
  illustrate motion filtering. The nearly static clip fails the 0.1 threshold;
  the clip showing movement of a hand-held tool passes.}
  \label{fig:example-motion}
\end{figure}

\begin{figure}[htbp]
  \centering
  \includegraphics[width=0.95\linewidth]{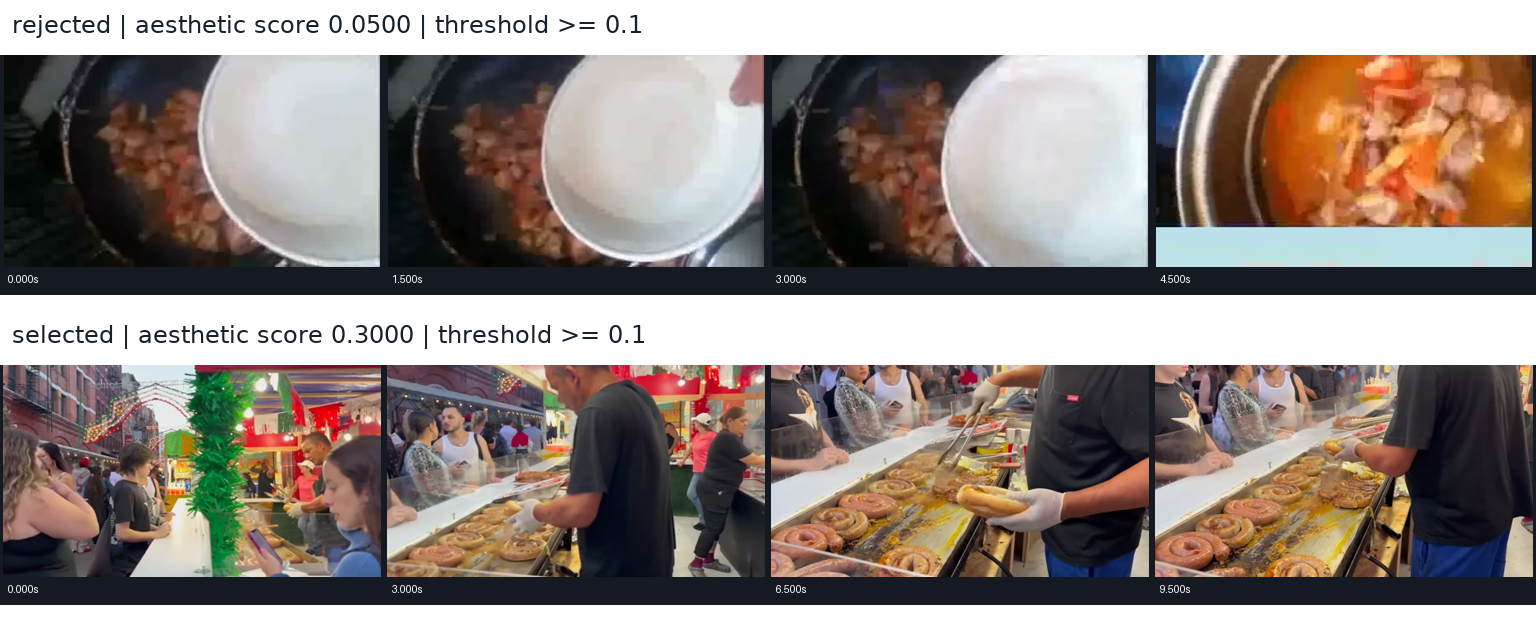}
  \caption{\textbf{Aesthetic filtering.} The clip scoring 0.05 is rejected,
  while the clip scoring 0.30 passes the 0.1 aesthetic threshold.}
  \label{fig:example-aesthetic}
\end{figure}

\begin{figure}[htbp]
  \centering
  \includegraphics[width=0.95\linewidth]{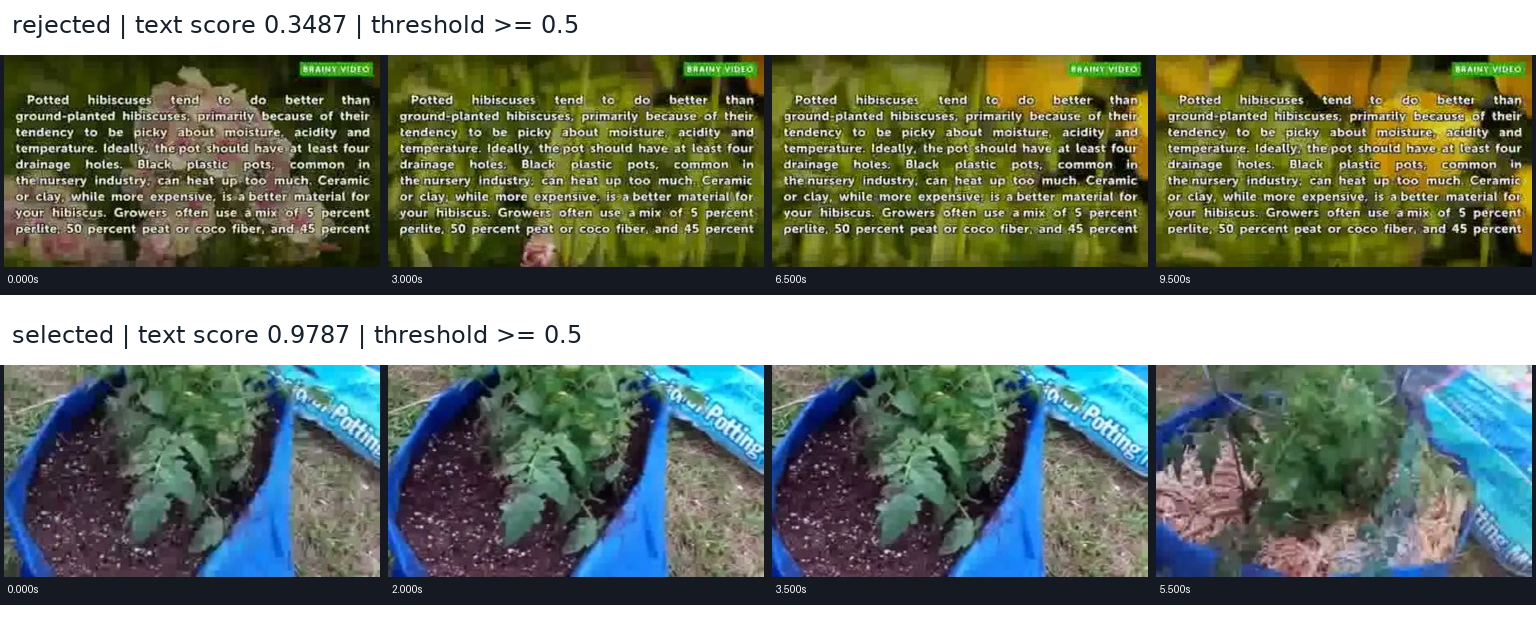}
  \caption{\textbf{Visible-text filtering.} Plant-related clips illustrate the
  0.5 threshold. A higher score indicates a smaller detected text area;
  the images show each clip's recorded score and selection outcome.}
  \label{fig:example-text}
\end{figure}

\begin{figure}[p]
  \centering
  \includegraphics[width=\linewidth,height=0.78\textheight,keepaspectratio]{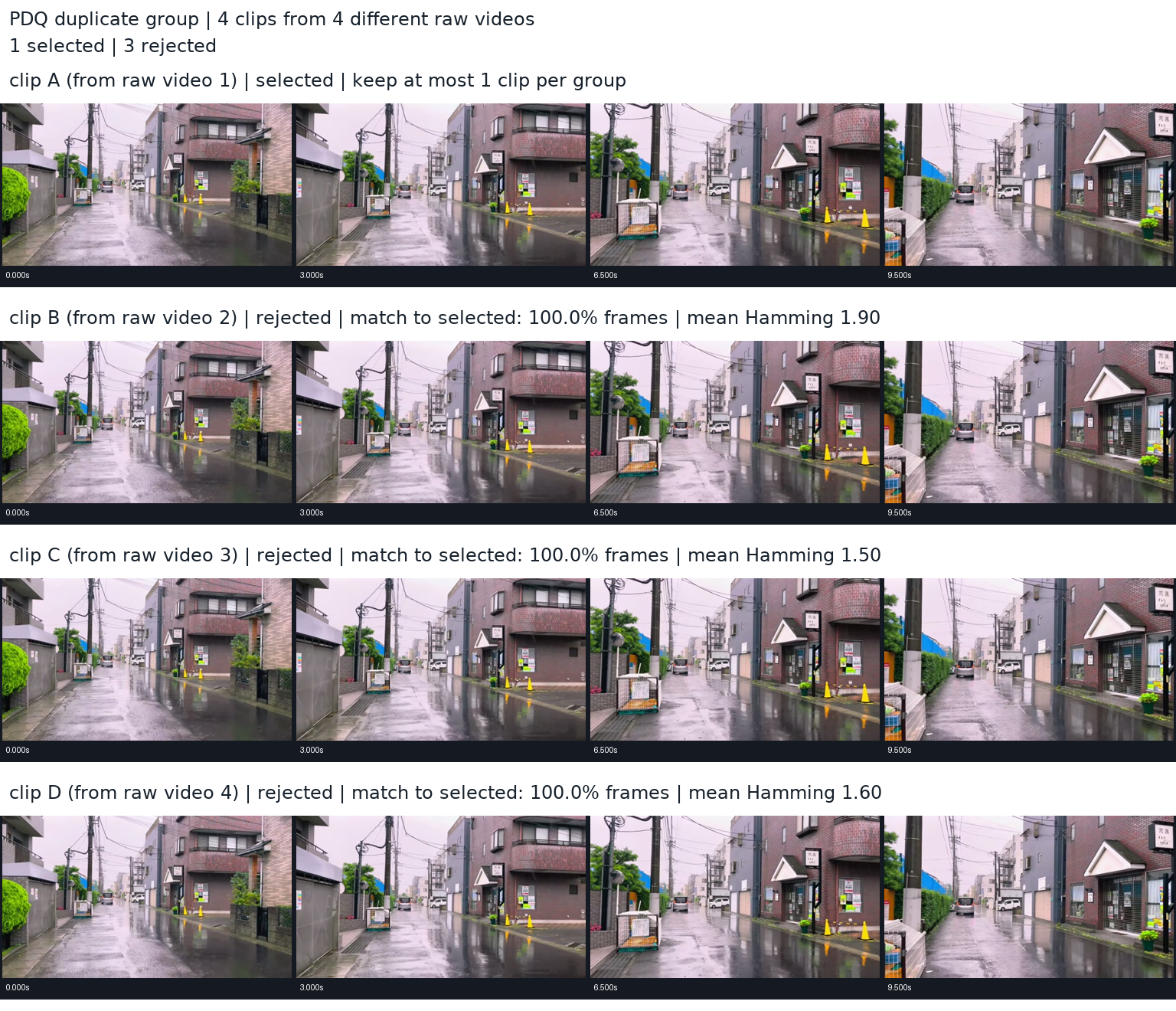}
  \caption{\textbf{Perceptual deduplication.} Four clips from different source
  videos show the same rainy street scene. All four pass the quality filters.
  PDQ identifies them as near-duplicates, and the recipe retains only 1 clip
  in the group.}
  \label{fig:example-pdq}
\end{figure}

\begin{figure}[p]
  \centering
  \includegraphics[width=\linewidth,height=0.82\textheight,keepaspectratio]{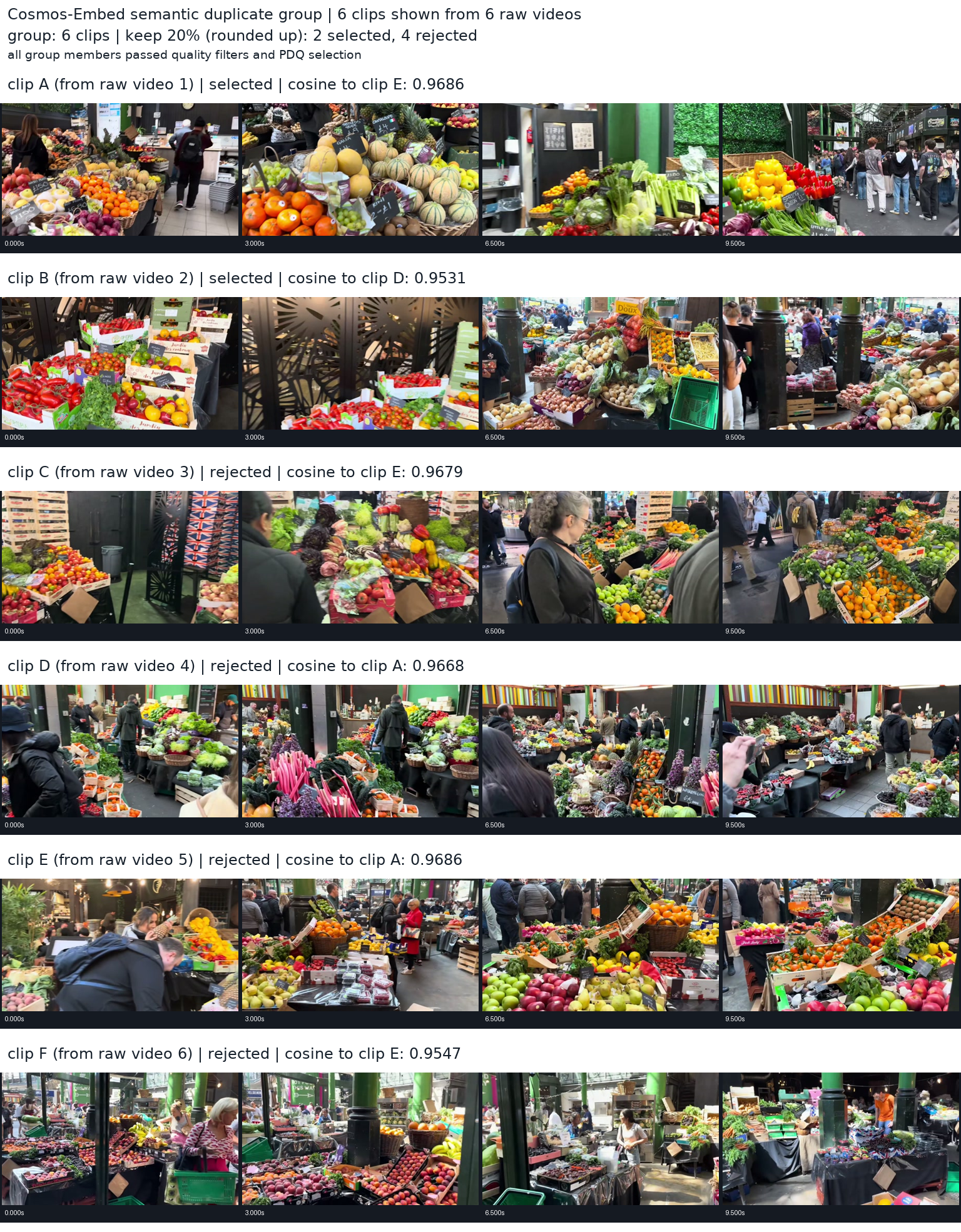}
  \caption{\textbf{Semantic deduplication.} Six market clips from different
  source videos pass the quality filters and PDQ selection. Cosmos-Embed groups
  them by semantic similarity. Keeping 20\% of the group, rounded up, retains
  2 clips and removes 4. Each similarity score compares the displayed clip
  with the clip named beside it.}
  \label{fig:example-cosmos}
\end{figure}

\Cref{fig:example-optical,fig:example-motion,fig:example-aesthetic,fig:example-text}
show clips that pass or fail the optical, motion, aesthetic, and visible-text filters.
\Cref{fig:example-pdq,fig:example-cosmos} show which clips are retained or removed
by PDQ and Cosmos-Embed deduplication. Selected examples belong to the Selected
and Mixed training sets; rejected examples belong to Rejected and Mixed.

\FloatBarrier
\subsection{Annotation Examples}
\label{app:annotation-examples}

\Cref{fig:sample-annotation-s1-prototype,fig:sample-annotation-r1-prototype}
show Camera and Tag annotations and all four caption levels for selected and rejected clips.
Teal highlights saved model choices; unchosen options are grey. Six Camera
fields and all seven Tag fields are shown; asterisks mark multi-label fields.

\begin{figure}[htbp]
  \centering
  \resizebox{!}{0.76\textheight}{\input{figures/annotation_s1.tex}}
  \caption{\textbf{Annotations for a selected clip.} A valley landscape is
  shown with its quality scores, selection outcome, Camera and Tag labels,
  and four caption levels, from a short summary to a detailed description.}
  \label{fig:sample-annotation-s1-prototype}
\end{figure}

\begin{figure}[p]
  \centering
  \resizebox{!}{0.88\textheight}{\input{figures/annotation_r1.tex}}
  \caption{\textbf{Annotations for a rejected clip.} An engine-repair clip
  fails the optical, motion, and aesthetic thresholds, yet retains Camera
  and Tag labels and all four caption levels for use in alternative recipes.}
  \label{fig:sample-annotation-r1-prototype}
\end{figure}

\FloatBarrier
\section{Limitations}
\label{app:limitations}

Our recipe study focuses on early pretraining with two model families;
recipe rankings may change with longer training or other architectures.
We evaluate filtering and deduplication as a combined selection policy,
without isolating the contribution of each operation.
The released annotations are automatically generated, and their semantic
accuracy has not yet been systematically assessed through human evaluation.

\end{document}